\documentclass[final,5p,times,twocolumn]{elsarticle}
\biboptions{sort&compress}
\usepackage{amssymb}
\usepackage{amsmath}
\usepackage{graphicx}
\usepackage{subcaption}
\usepackage{booktabs}
\usepackage{multirow}
\usepackage{makecell}
\usepackage{array}
\usepackage{float}
\usepackage{tabularx}
\usepackage{booktabs}
\usepackage{graphicx}
\usepackage{algorithm}
\usepackage{algpseudocode}
\usepackage{adjustbox}
\usepackage{xcolor}

\usepackage{booktabs}
\usepackage{etoolbox}
\usepackage{tabularx}
\usepackage{array}

\newcommand{\unifiedtablestyle}{%
    \centering
    \footnotesize
    \renewcommand{\arraystretch}{1.08}%
    \setlength{\tabcolsep}{4pt}%
}

\AtBeginEnvironment{table}{\unifiedtablestyle}
\AtBeginEnvironment{table*}{\unifiedtablestyle}

\newcommand{\placeholdercell}[1]{%
  \fbox{%
    \begin{minipage}[c][1.55cm][c]{0.94\linewidth}
      \centering
      \scriptsize #1\\[0.2em]
      \textit{placeholder}
    \end{minipage}%
  }%
}

\newcommand{\qualcell}[2]{%
  \IfFileExists{#1}{%
    \includegraphics[width=0.94\linewidth]{#1}%
  }{%
    \placeholdercell{#2}%
  }%
}

\definecolor{mycitecolor}{RGB}{50,150,180}

\usepackage[
    colorlinks=true,
    linkcolor=black,
    citecolor=mycitecolor,
    urlcolor=blue
]{hyperref}

\usepackage{stfloats}

\journal{Information Fusion}

\begin{document}

\begin{frontmatter}


\title{Robust 3D Reconstruction from Multi-View Optical Satellite Imagery via Reliability-Aware Height-Evidence Fusion in Gaussian Splatting}


\author[inst1]{Jie Yang}
\ead{yangjie1@whu.edu.cn}

\author[inst1,inst2]{Yingdong Pi\corref{cor1}}
\ead{pyd\_imars@whu.edu.cn}

\author[inst1]{Qiyan Luo}
\ead{luoqy26@whu.edu.cn}

\author[inst3]{Xiaoyu Wang}
\ead{xiaoyu.wang@whu.edu.cn}

\author[inst1]{Lekang Wen}
\ead{wenlk3@whu.edu.cn}

\author[inst1,inst2]{Mi Wang}
\ead{wangmi@whu.edu.cn}

\cortext[cor1]{Corresponding author.}

\affiliation[inst1]{
    organization={State Key Laboratory of Information Engineering in Surveying, Mapping and Remote Sensing, Wuhan University},
    addressline={129 Luoyu Road},
    city={Wuhan},
    postcode={430079},
    country={China}
}

\affiliation[inst2]{
    organization={Hubei Luojia Laboratory},
    addressline={129 Luoyu Road},
    city={Wuhan},
    postcode={430079},
    country={China}
}

\affiliation[inst3]{
    organization={School of Computer Science, Wuhan University},
    addressline={129 Luoyu Road},
    city={Wuhan},
    postcode={430079},
    country={China}
}

%


\begin{abstract}
Robust 3D reconstruction from multi-view optical satellite imagery requires fusing complementary but sometimes conflicting geometric evidence. Digital surface models (DSMs) are the primary elevation representations for satellite-based 3D reconstruction, making reliable height estimation essential. However, in a Gaussian scene representation jointly optimized from multiple views, Gaussian responses at different elevations can support competing height hypotheses at the same rendered location, while conventional alpha-weighted elevation aggregation may produce intermediate elevations that do not correspond to physical surfaces. To address this challenge, we formulate DSM reconstruction as a reliability-aware height-hypothesis fusion problem and propose HLC-GS, a reliability-aware Height-Layer Consistency Gaussian Splatting framework for multi-view satellite 3D reconstruction. HLC-GS organizes projected Gaussian responses into candidate height hypotheses and evaluates their relative support using layer competition and Gaussian footprint support. A continuous height-layer risk map guides dominant-layer reliability correction and secondary-layer suppression during optimization. The proposed training strategy regulates conflicting Gaussian responses within the shared representation to improve the reliability of reconstructed surface elevations. Experiments on seven scenes from the DFC2019 and IARPA2016 datasets demonstrate improved DSM reconstruction accuracy. Compared with EOGS, HLC-GS reduces the average DSM MAE from 1.46~m to 1.18~m and RMSE from 2.78~m to 2.58~m, while increasing PAG$_{2.5}$ from 86.09\% to 88.61\%, with comparable computational cost.
\end{abstract}





\end{frontmatter}


\section{Introduction}
\label{sec:introduction}

Robust 3D reconstruction from multi-view optical satellite imagery is essential for large-scale Earth observation and geospatial analysis. By exploiting complementary observations acquired from different viewing geometries, satellite remote sensing enables wide-area and cost-effective reconstruction of scene geometry~\citep{amadei2025s2p,yao2025magiccity}. The recovered geometry is commonly represented as a Digital Surface Model (DSM), which describes the elevation distribution of visible natural and man-made surfaces. Therefore, achieving accurate and robust 3D reconstruction by effectively integrating multi-view optical satellite observations remains a fundamental problem in photogrammetry and Earth observation~\citep{li20243d,lu2021satmvs}.

Recovering reliable 3D geometry from multi-view satellite imagery remains challenging because observations acquired from different viewpoints and times provide complementary yet often incomplete and inconsistent geometric information. Traditional photogrammetric pipelines and multi-view stereo (MVS) methods~\citep{d2012dense,yao2018mvsnet,gong2019dsm,qin2019automated,qin2019critical} integrate these observations mainly through cross-image correspondence and geometric matching, but remain sensitive to weak textures, occlusions, shadows, and radiometric variations caused by multi-temporal acquisition. Neural Radiance Field (NeRF)-based methods~\citep{mildenhall2021nerf} instead jointly optimize a continuous scene representation from multiple observations, reducing the dependence on explicit dense matching. However, their volumetric optimization is computationally demanding, and surface geometry is typically recovered indirectly from the learned scene representation, which can limit geometric accuracy under photometric ambiguity~\citep{mari2022sat,mari2023multi,qu2023sat}.

\begin{figure*}[t]
\centering
\includegraphics[width=0.9\textwidth]{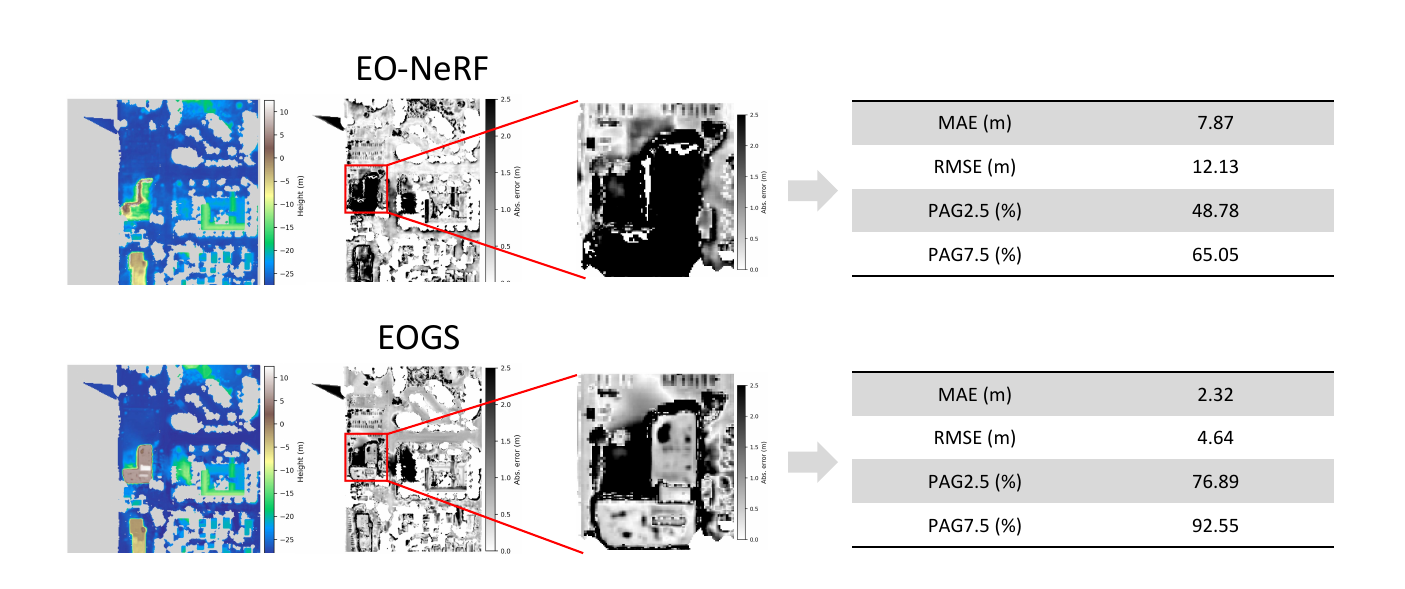}
\caption{Motivation of HLC-GS using a representative L-shaped building region from JAX\_260. Although EOGS~\citep{aira2025gaussian} improves the reconstruction over EO-NeRF~\citep{mari2023multi}, noticeable elevation errors remain near building boundaries and roof--ground transitions, where Gaussian responses associated with vertically separated surfaces can contribute simultaneously. In the highlighted ROI, EOGS yields an MAE of \(2.32\,\mathrm{m}\) and an RMSE of \(4.64\,\mathrm{m}\). These localized errors motivate explicit reasoning over competing height hypotheses to reduce height-layer mixing during DSM reconstruction. All absolute-error maps use the same range, with white and black indicating smaller and larger errors, respectively.}
\label{fig:motivation}
\end{figure*}

The emergence of 3D Gaussian Splatting (3DGS)~\citep{kerbl20233d} provides an efficient explicit representation for multi-view 3D reconstruction, enabling fast differentiable rendering and direct optimization of Gaussian geometry. However, photometric optimization alone does not explicitly enforce geometric surface consistency, and the recovered geometry may remain unreliable in regions with depth discontinuities or incomplete observations. Moreover, the original 3DGS formulation is not designed for the imaging geometry of optical satellite observations. Surface-aware methods such as 2DGS~\citep{huang20242d} improve geometric surface modeling by representing geometry with oriented Gaussian surfels and introducing depth-distortion and normal-consistency constraints, thereby promoting more compact geometry and improved local surface consistency. These methods primarily focus on regularizing surface structure, but do not directly assess the competition among multiple elevation hypotheses during satellite DSM estimation. Consequently, when vertically separated Gaussian responses coexist along the same viewing direction, the ambiguity among competing elevation hypotheses may remain unresolved. For satellite reconstruction, EOGS~\citep{aira2025gaussian} further adapts Gaussian Splatting to multi-view satellite observations and enables efficient recovery of surface elevations from the optimized Gaussian representation. However, its elevation estimation still relies on alpha-weighted aggregation and does not explicitly model competing height layers or their reliability. Consequently, Gaussian responses associated with different elevations may still be blended during aggregation, producing intermediate elevations that do not correspond to any physical surface. Fig.~\ref{fig:motivation} provides a representative example: although EOGS improves reconstruction over EO-NeRF, clear residual errors remain near building boundaries and roof--ground transitions, where competing height responses are particularly pronounced. This unresolved ambiguity indicates that accurate satellite 3D reconstruction requires more than generic surface regularization or weighted elevation aggregation. Since multiple views jointly constrain a shared Gaussian representation, different Gaussian responses may provide competing explanations of the surface geometry at the same location. Therefore, a key challenge is to determine the reliability of these competing height hypotheses and adapt their contributions during reconstruction.

To address this problem, we propose HLC-GS, a reliability-aware Height-Layer Consistency Gaussian Splatting framework that models and regulates competing height hypotheses within a shared Gaussian representation for multi-view optical satellite imagery. The views jointly constrain a shared Gaussian scene, and projected responses from this scene are organized at each rendered location into candidate height layers. HLC-GS assesses the relative support of these hypotheses by combining layer competition and Gaussian footprint support. A continuous height-layer risk map identifies ambiguous or unreliable configurations, while dominant-layer reliability correction (DRC) and secondary-layer suppression (SLS) adapt the training-time influence of unreliable dominant responses and interfering secondary responses. Therefore, the proposed reliability modeling improves DSM reconstruction by regulating the learned representation rather than introducing an additional post-processing selection strategy.

The main contributions of this work are summarized as follows:
\begin{itemize}
\item We formulate multi-view satellite 3D reconstruction as a reliability-aware fusion problem over competing height hypotheses within a shared Gaussian representation. We identify height-layer mixing in Gaussian-based satellite DSM reconstruction, where alpha blending of responses from different elevation layers produces non-physical intermediate elevations.

\item We propose HLC-GS, a reliability-aware Height-Layer Consistency Gaussian Splatting framework that evaluates the reliability of competing height hypotheses and regulates conflicting Gaussian responses during optimization. A risk-guided optimization strategy, including dominant-layer reliability correction and secondary-layer suppression, is introduced to adapt the contributions of unreliable height responses within the shared Gaussian representation.

\item We validate HLC-GS on seven scenes from the DFC2019~\cite{bosch2019semantic} and IARPA2016~\cite{bosch2016multiple} benchmarks through comprehensive comparisons, component ablations, and reliability analyses, demonstrating the effectiveness of reliability-guided regulation of competing height hypotheses for improving DSM reconstruction accuracy.
\end{itemize}

The remainder of this paper is organized as follows. Section~\ref{sec:related_work} reviews related work on multi-view satellite 3D reconstruction, neural implicit reconstruction with geometric constraints, geometric regularization and surface modeling in Gaussian Splatting, and Gaussian Splatting for satellite scene reconstruction. Section~\ref{sec:method} presents the proposed HLC-GS framework. Section~\ref{sec:experiments} describes the datasets, evaluation protocols, implementation details, and comparative methods. Section~\ref{sec:results} presents the quantitative and qualitative results together with comprehensive analyses. Finally, Section~\ref{sec:conclusion} concludes the paper.

\section{Related Work}
\label{sec:related_work}

In this section, we review related studies on multi-view satellite 3D reconstruction, covering traditional photogrammetric pipelines, neural implicit representations, geometric regularization and surface modeling in Gaussian Splatting, and recent Gaussian Splatting methods designed for satellite scene reconstruction.

\subsection{Multi-view Satellite 3D Reconstruction}
\label{subsec:traditional_dsm}

Multi-view satellite 3D reconstruction has traditionally relied on photogrammetric stereo matching, multi-view depth estimation, and geometric aggregation. Representative systems such as S2P, the Ames Stereo Pipeline (ASP), and MicMac explicitly model satellite imaging geometry and combine dense correspondence estimation with point-cloud or DSM generation~\cite{de2014automatic,shean2016automated,rupnik2017micmac}. Dense matching algorithms, including Semi-Global Matching (SGM) and More Global Matching (MGM), remain widely used in these pipelines~\citep{hirschmuller2008stereo,facciolo2015mgm}. When multi-temporal observations are available, integrating images acquired from different viewpoints and dates can improve geometric completeness and robustness, but reconstruction quality remains sensitive to stereo-pair selection, illumination variation, occlusion, and temporal changes~\citep{qin2019automated,qin2019critical}.

Because geometric evidence from different observations is not equally reliable, several studies further introduce confidence- or uncertainty-aware strategies during multi-view depth and surface aggregation~\citep{qin2022uncertainty}. These methods estimate the quality of individual depth hypotheses and assign them different contributions during fusion, thereby reducing the influence of unreliable observations. Such approaches demonstrate the importance of reliability-aware integration of multi-view geometric information. However, reliability is typically assessed after correspondence or depth estimation, at the depth-map, point-cloud, or DSM level. In contrast, HLC-GS reasons about reliability within the optimized Gaussian representation, where competing height hypotheses are identified and selectively regulated before the final surface elevation is generated.

\subsection{Neural Implicit Reconstruction with Geometric Constraints}
\label{subsec:satellite_nerf}

Neural radiance fields represent a scene as a continuous volumetric field and jointly optimize radiance and density through differentiable rendering~\citep{mildenhall2021nerf}. For satellite imagery, several studies adapt this formulation to satellite imaging geometry, illumination variation, and multi-temporal observations. S-NeRF explicitly models solar illumination and cast shadows~\citep{derksen2021shadow}, while Sat-NeRF incorporates RPC-based camera geometry, shadow-aware rendering, and uncertainty modeling for robust multi-date reconstruction~\citep{mari2022sat}. EO-NeRF further introduces geometry-consistent shadow modeling to improve elevation recovery~\citep{mari2023multi}, whereas SatelliteRF reduces computational cost through compact feature encodings~\citep{zhou2024satelliterf}. These methods demonstrate the potential of neural radiance fields for satellite scene reconstruction, but surface geometry is still recovered primarily from an implicitly optimized volumetric representation.

To improve geometric fidelity, subsequent neural implicit methods introduce stronger surface priors and explicit multi-view constraints. NeuS~\citep{wang2021neus} represents geometry as the zero level set of a signed distance function (SDF) and couples surface reconstruction with volume rendering, while Geo-NeuS~\citep{fu2022geo} incorporates multi-view geometric consistency to improve surface estimation. Neuralangelo~\citep{li2023neuralangelo} further enhances high-fidelity reconstruction through multi-resolution neural representations and geometric regularization. Similar developments have also emerged in satellite reconstruction. Sat-Mesh combines an SDF representation with multi-view stereo patch-matching constraints~\citep{qu2023sat}, while GC-NeRF introduces occupancy-guided sampling, vertical scene regularization, and multi-view DSM fusion~\citep{wan2024constraining}. These studies show that incorporating complementary geometric evidence can reduce ambiguities that cannot be resolved by photometric optimization alone.

However, these geometric constraints are primarily imposed on implicit density fields, SDFs, depth estimates, or reconstructed surfaces. HLC-GS instead performs reliability reasoning directly over competing height hypotheses induced by projected Gaussian primitives within an explicit Gaussian representation. By evaluating the relative support and reliability of these hypotheses during optimization, ambiguous height responses can be selectively regulated before the final surface elevation is generated.

\subsection{Geometric Regularization and Surface Modeling in Gaussian Splatting}
\label{subsec:geometry_aware_gs}

Although 3D Gaussian Splatting enables efficient differentiable rendering, photometric supervision alone does not guarantee a geometrically coherent surface representation. Recent studies have therefore introduced surface-oriented representations and explicit geometric constraints to improve Gaussian-based reconstruction. SuGaR~\citep{guedon2024sugar} encourages Gaussian primitives to align with the underlying surface for accurate mesh extraction, while 2D Gaussian Splatting (2DGS)~\citep{huang20242d} represents scene geometry using oriented Gaussian disks and introduces depth-distortion and normal-consistency regularization to promote compact and locally coherent surfaces. PGSR~\citep{chen2024pgsr} further incorporates planar priors, unbiased depth rendering, and multi-view geometric constraints to improve surface reconstruction. These methods demonstrate that explicitly regularizing Gaussian geometry can substantially improve surface quality beyond purely photometric optimization.

Multi-view geometric information has also been incorporated directly into Gaussian optimization. MVG-Splatting~\citep{li2026mvg} introduces multi-view geometric consistency for depth--normal refinement and adaptive densification, exploiting cross-view observations to improve geometric stability during reconstruction. Overall, these approaches mainly enhance the geometric organization, surface regularity, and cross-view consistency of Gaussian representations. However, improved surface coherence does not necessarily resolve which surface should be trusted when multiple vertically separated Gaussian responses contribute to the same rendered location. HLC-GS addresses this complementary problem by explicitly organizing these responses into competing height hypotheses, evaluating their relative support and reliability, and selectively regulating unreliable dominant and interfering secondary responses during optimization. In this way, HLC-GS targets ambiguity in height evidence rather than relying solely on generic surface regularization.

\subsection{Gaussian Splatting for Satellite Scene Reconstruction}
\label{subsec:satellite_3dgs}

The efficiency and explicit scene representation of 3D Gaussian Splatting (3DGS) have motivated growing interest in satellite 3D reconstruction. EOGS~\citep{aira2025gaussian} adapts 3DGS to satellite photogrammetry through affine camera approximation, shadow modeling, and radiometric correction, providing an efficient framework for reconstructing surface geometry from multi-view satellite imagery. Subsequent studies extend satellite Gaussian reconstruction from different perspectives. SatGS~\citep{bai2025satgs} and SA-GS~\citep{xu2026sa} improve robustness to multi-temporal appearance and seasonal variations through appearance-adaptive modeling and satellite-view augmentation. SkySplat~\citep{huang2025skysplat} targets sparse multi-temporal observations using RPC-aware modeling and multi-view consistency aggregation, while RPC-GS~\citep{wagner2026rpc} directly incorporates the RPC camera model into Gaussian rasterization to reduce geometric errors caused by simplified projection models.

Recent studies have also placed increasing emphasis on surface geometry and the reliability of geometric information. SatSplat~\citep{song2026satsplat} extends 2D Gaussian Splatting to satellite photogrammetry and jointly refines camera parameters and surface geometry, whereas SatSurfGS~\citep{chen2026satsurfgs} models spatially varying geometric reliability and adaptively combines monocular priors with multi-view matching evidence for sparse-view surface reconstruction. Collectively, these methods improve satellite Gaussian reconstruction through more accurate imaging models, stronger appearance adaptation, enhanced surface representations, and more effective use of multi-view geometric information. However, the ambiguity arising from competing height responses within the Gaussian representation remains insufficiently explored. HLC-GS addresses this complementary problem by explicitly organizing vertically separated responses into competing height hypotheses, evaluating their relative support and reliability, and selectively regulating unreliable dominant and interfering secondary responses during optimization. In this way, height ambiguity that would otherwise propagate into weighted surface-elevation aggregation is addressed directly within the reconstruction process.

\begin{figure*}[t]
\centering
\includegraphics[width=0.95\textwidth]{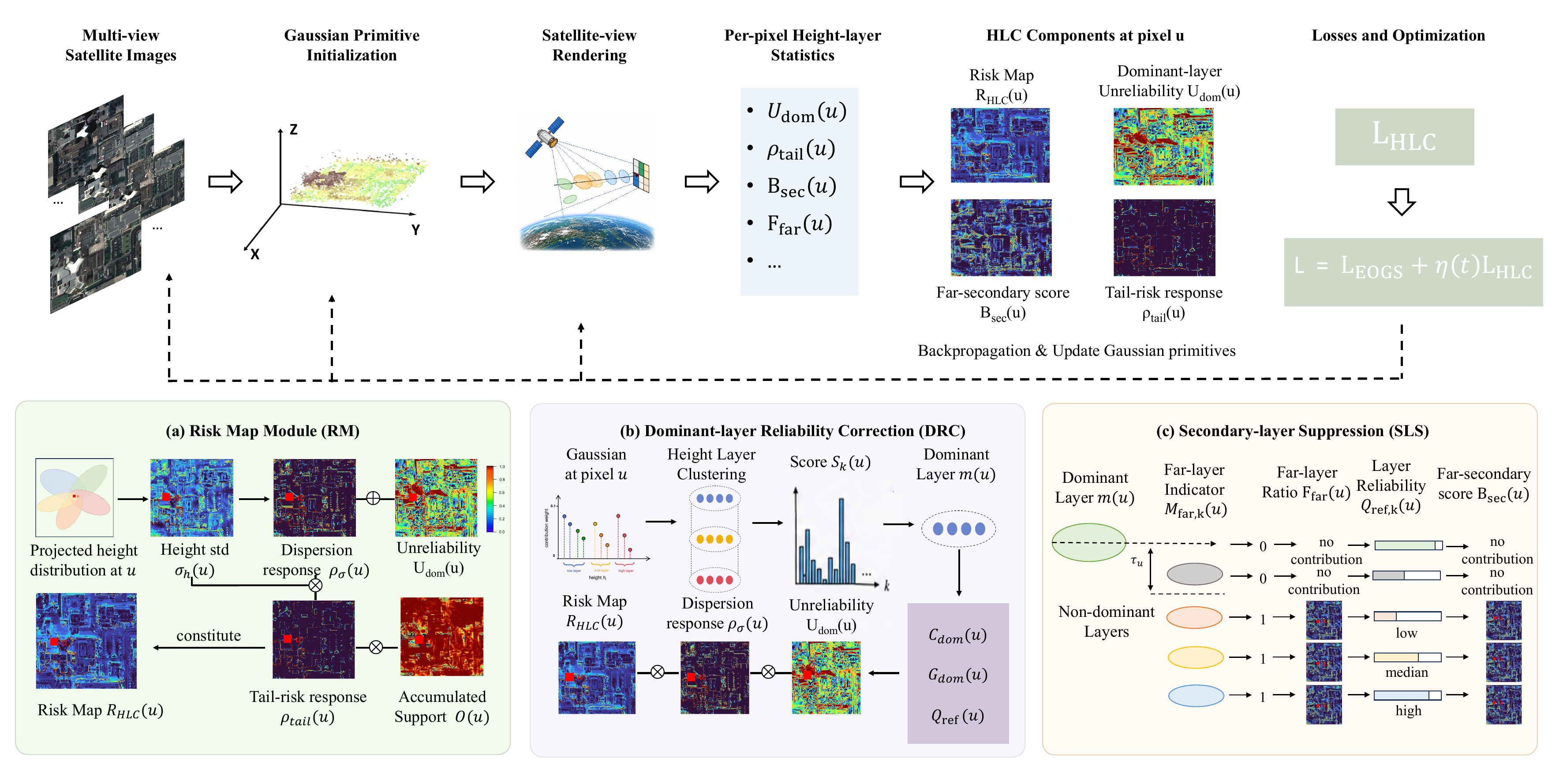}
\caption{Overview of reliability-aware height-hypothesis regulation in HLC-GS. Multi-view satellite observations jointly constrain a shared Gaussian scene representation. Projected Gaussian responses at each pixel are organized into candidate height hypotheses with potentially conflicting support. The risk map (RM) characterizes the ambiguity of these hypotheses, while dominant-layer reliability correction (DRC) and secondary-layer suppression (SLS) regulate the corresponding Gaussian responses during optimization.}
\label{fig:framework}
\end{figure*}

\section{Methodology}
\label{sec:method}

\subsection{Preliminaries and Problem Formulation}
\label{sec:preliminaries}

Given a set of multi-view optical satellite images with known imaging models, our goal is to reconstruct a DSM representing the surface elevation of the observed scene. Different views provide complementary but potentially inconsistent geometric constraints due to variations in visibility, illumination, and acquisition time. HLC-GS represents the scene using a shared set of anisotropic 3D Gaussian primitives and optimizes them through differentiable rasterization adapted to satellite imaging geometry. We denote the Gaussian representation as $G=\{g_i\}$, where each primitive is parameterized by its 3D center, covariance, opacity, and appearance attributes. The shared representation accumulates geometric evidence from multiple views and is subsequently used for explicit elevation rendering. 

During rasterization, Gaussian primitives contributing to the same rendered pixel are composited in front-to-back order. Omitting the view index for simplicity, let \(u\in\Omega\) denote a valid pixel and let \(\mathcal{G}(u)\) denote the ordered set of Gaussian primitives contributing to \(u\). Following EOGS, surface elevation is rendered by replacing the color attribute of each Gaussian with the real-world altitude of its center. Let \(h_i\) denote the altitude of the \(i\)-th Gaussian and \(\omega_i(u)\) its alpha-compositing weight at pixel \(u\). The rendered elevation is given by
\begin{equation}
H(u)=
\frac{
\sum_{i\in\mathcal{G}(u)}\omega_i(u)h_i
}{
\sum_{i\in\mathcal{G}(u)}\omega_i(u)+\epsilon
},
\label{eq:weighted_height}
\end{equation}
where \(\epsilon\) is a small constant for numerical stability. The resulting altitude map is combined with the image-coordinate grid, back-projected to geospatial coordinates through the satellite imaging model, and finally rasterized onto the target DSM grid.

Although Eq.~(1) provides a differentiable elevation estimate, it aggregates all Gaussian responses along the rendering ray according to their compositing weights without distinguishing whether they belong to the same physical surface. As introduced in Sec.~\ref{sec:introduction}, this can lead to \emph{height-layer mixing} when vertically separated responses coexist at the same rendered location, producing an intermediate elevation between competing height hypotheses. HLC-GS therefore explicitly organizes these responses into candidate height layers and reasons about their ambiguity and reliability before applying reliability-aware height-layer regulation.

The central problem is therefore to distinguish meaningful height hypotheses from conflicting responses and to determine whether the locally dominant hypothesis is sufficiently reliable to represent the surface. HLC-GS addresses this problem by introducing explicit height-layer reasoning into Gaussian optimization. Projected Gaussian responses are first organized into candidate height layers, from which local height dispersion and layer competition are characterized. Complementary cues from dominant-layer strength and Gaussian footprint support are then used to assess dominant-hypothesis reliability. Based on these statistics, a continuous risk map \(R_{\mathrm{HLC}}(u)\) identifies ambiguous or unreliable height configurations, while dominant-layer reliability correction and secondary-layer suppression selectively regulate unreliable dominant and distant competing responses. The resulting optimization reduces height-layer conflicts without imposing a globally single-layer surface assumption. An overview of the framework is shown in Fig.~\ref{fig:framework}.

\subsection{Risk Map Module (RM)}
\label{sec:risk_map}

The risk map module characterizes ambiguous and unreliable height configurations and provides spatially adaptive weights for subsequent HLC optimization. Rather than applying uniform reliability-aware regulation to all rendered locations, RM emphasizes regions exhibiting large vertical dispersion or insufficiently reliable dominant height hypotheses.

For a valid pixel \(u\in\Omega\), let \(\mathcal{G}(u)\) denote the front-to-back ordered Gaussian primitives contributing to \(u\), with altitude \(h_i\), pixel opacity \(\alpha_i(u)\), and alpha-compositing weight \(\omega_i(u)\). We first quantify the vertical dispersion of the contributing responses using the weighted altitude standard deviation
\begin{equation}
\sigma_h(u)=
\left(
\frac{\sum_{i\in\mathcal{G}(u)}\omega_i(u)h_i^2}
{\sum_{i\in\mathcal{G}(u)}\omega_i(u)+\epsilon}
-
\left(
\frac{\sum_{i\in\mathcal{G}(u)}\omega_i(u)h_i}
{\sum_{i\in\mathcal{G}(u)}\omega_i(u)+\epsilon}
\right)^2
\right)^{1/2}.
\label{eq:height_std}
\end{equation}
A large \(\sigma_h(u)\) indicates that Gaussian responses with substantially different elevations contribute to the same rendered location.

Because the magnitude of height dispersion varies across scenes and views, we normalize $\sigma_h(u)$ according to its empirical distribution over valid pixels. We define a dispersion response $\rho_{\sigma}(u)$ and an upper-tail response $\rho_{\mathrm{tail}}(u)$ as
\begin{equation}
\begin{aligned}
\rho_{\sigma}(u)
&=
\operatorname{clip}\!\left(
\frac{\sigma_h(u)-Q_{0.75}(\sigma_h)}
{\max\{Q_{0.95}(\sigma_h)-Q_{0.75}(\sigma_h),10^{-4}\}},
0,1
\right),\\
\rho_{\mathrm{tail}}(u)
&=
\operatorname{clip}\!\left(
\frac{\sigma_h(u)-Q_{0.90}(\sigma_h)}
{\max\{Q_{0.99}(\sigma_h)-Q_{0.90}(\sigma_h),10^{-4}\}},
0,1
\right),
\end{aligned}
\label{eq:rho_dispersion}
\end{equation}
where $Q_p(\sigma_h)$ denotes the $p$-th quantile of the height-dispersion distribution over valid pixels in the current rendered view. The response $\rho_{\sigma}(u)$ characterizes generally elevated vertical dispersion, whereas $\rho_{\mathrm{tail}}(u)$ specifically emphasizes pixels in the extreme upper tail of the dispersion distribution. The former provides the primary dispersion cue for risk estimation, while the latter further increases the emphasis on severe height-layer ambiguity.

Height dispersion alone cannot determine which response should represent the local surface. We therefore organize the contributing Gaussian responses into candidate height layers according to their elevation proximity. Responses are traversed in front-to-back compositing order, and those with sufficiently similar elevations are grouped into the same candidate layer according to a GSD-aware height threshold \(\tau_h=\operatorname{clip}(2\,\mathrm{GSD},0.5\,\mathrm{m},1.5\,\mathrm{m})\). This construction separates vertically distinct response groups while adapting the layer resolution to the spatial resolution of the satellite imagery.

For each candidate layer \(\mathcal{C}_k(u)\), we compute
\begin{equation}
\begin{aligned}
W_k(u)
&=
\sum_{i\in\mathcal{C}_k(u)}\omega_i(u),
\qquad
\bar h_k(u)
=
\frac{
\sum_{i\in\mathcal{C}_k(u)}\omega_i(u)h_i
}{
W_k(u)+\epsilon
},\\
S_k(u)
&=
W_k(u)
\sqrt{Q_{\mathrm{foot},k}(u)+\epsilon},
\end{aligned}
\label{eq:layer_statistics}
\end{equation}
where \(W_k(u)\) denotes the accumulated compositing contribution, \(\bar h_k(u)\) the representative elevation, and \(Q_{\mathrm{foot},k}(u)\) the Gaussian footprint support of the \(k\)-th layer. The score \(S_k(u)\) therefore favors height hypotheses supported by both strong compositing contribution and stable footprint evidence. The dominant layer is selected as \(m(u)=\arg\max_k S_k(u)\), with representative elevation \(H_{\mathrm{dom}}(u)=\bar h_{m(u)}(u)\).

The strength and separability of the dominant hypothesis are characterized by
\begin{equation}
C_{\mathrm{dom}}(u)=
\frac{S_{m(u)}(u)}
{\sum_k S_k(u)+\epsilon},
\qquad
G_{\mathrm{dom}}(u)=
\frac{S_{m(u)}(u)-S_{(2)}(u)}
{S_{m(u)}(u)+\epsilon},
\label{eq:dom_conf_gap}
\end{equation}
where \(S_{(2)}(u)\) denotes the second-largest layer score. \(C_{\mathrm{dom}}(u)\) measures the relative support of the dominant hypothesis, whereas \(G_{\mathrm{dom}}(u)\) measures its separation from the strongest competing hypothesis.

A dominant hypothesis may still be unreliable even when it clearly wins the local competition. We therefore assess its reliability by combining complementary cues from layer dominance and Gaussian footprint support:
\begin{equation}
\begin{aligned}
\bar Q_{\mathrm{foot}}(u)
&=
\frac{
\sum_k\sum_{i\in\mathcal C_k(u)}
\omega_i(u)\alpha_i(u)
}{
\sum_{i\in\mathcal G(u)}
\omega_i(u)+\epsilon
},\\
Q_{\mathrm{rel}}(u)
&=
C_{\mathrm{dom}}(u)\,
\bar Q_{\mathrm{foot}}(u),\\
U_{\mathrm{dom}}(u)
&=
\phi(C_{\mathrm{dom}}(u);\tau_c)
\phi(G_{\mathrm{dom}}(u);\tau_g)
[\tau_u-Q_{\mathrm{rel}}(u)]_+ .
\end{aligned}
\label{eq:dominant_reliability}
\end{equation}
Here, $\bar Q_{\mathrm{foot}}(u)$ summarizes the aggregate footprint support of the contributing Gaussian responses. We define the soft activation as $\phi(x;\tau)=[x-\tau]_+/(1-\tau)$, with $\tau_c=0.45$, $\tau_g=0.15$, and $\tau_u=0.55$. A large $Q_{\mathrm{rel}}(u)$ indicates that the dominant hypothesis is consistently supported by both layer dominance and Gaussian footprint evidence, whereas $U_{\mathrm{dom}}(u)$ assigns a larger response to dominant hypotheses that appear strong in local competition but remain insufficiently reliable.

Finally, the height-layer risk map is defined as
\begin{equation}
R_{\mathrm{HLC}}(u)=
\left(
\rho_{\sigma}(u)+U_{\mathrm{dom}}(u)
\right)
O(u)
\left(
1+3\rho_{\mathrm{tail}}(u)
\right),
\label{eq:hlc_risk}
\end{equation}
where \(O(u)=\sum_i\omega_i(u)\) denotes the accumulated compositing support. Consequently, \(R_{\mathrm{HLC}}(u)\) emphasizes well-supported locations exhibiting substantial vertical dispersion or insufficient dominant-hypothesis reliability, while further increasing the weight of extreme dispersion cases.

\begin{figure}[t]
\centering
\includegraphics[width=0.47\linewidth,height=3.2cm]{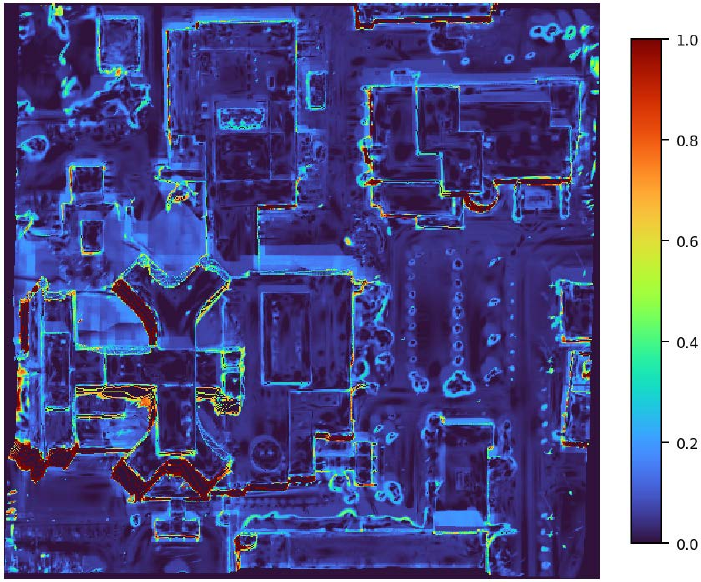}
\hfill
\includegraphics[width=0.47\linewidth,height=3.2cm]{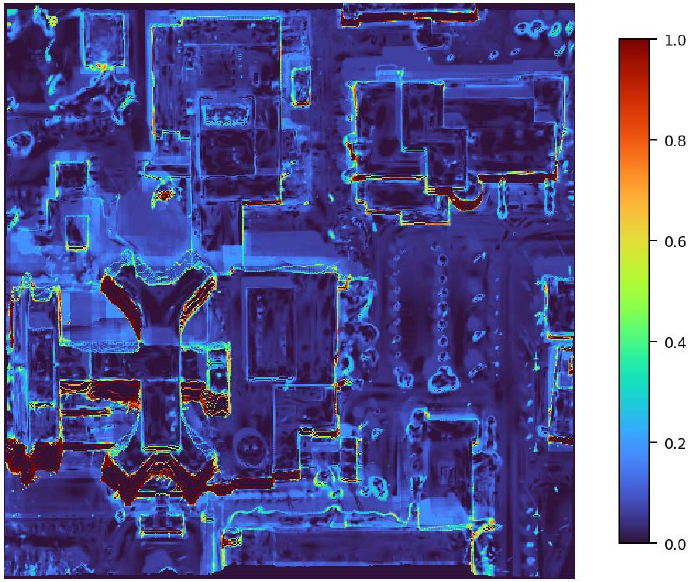}
\caption{
Qualitative visualization of the computed HLC risk maps on representative test views under the HLC-GS setting.
Warmer colors indicate higher risk values.
High-risk responses are mainly concentrated around building boundaries, roof--ground transitions, and other geometrically complex regions.
}
\label{fig:risk_map_vis}
\end{figure}

Representative risk maps are shown in Fig.~\ref{fig:risk_map_vis}. High-risk responses are mainly observed around building boundaries, roof--ground transitions, and other regions with pronounced height discontinuities. The ability of the proposed risk map to localize regions that subsequently benefit from HLC optimization is quantitatively evaluated in Section~\ref{subsubsec:risk_localization}.

\subsection{Dominant-layer Reliability Correction (DRC)}
\label{sec:dominant_layer}

The dominant-layer reliability correction (DRC) module targets configurations in which a height hypothesis is selected as the dominant layer but remains insufficiently reliable according to the combined geometric cues. As defined in Section~\ref{sec:risk_map}, \(U_{\mathrm{dom}}(u)\) becomes large when the selected hypothesis exhibits sufficient dominance and separability while its reliability \(Q_{\mathrm{rel}}(u)\) remains below the required level. This distinction prevents competitive dominance alone from being treated as evidence of a reliable surface estimate.

DRC regularizes locations where unreliable dominance coincides with substantial vertical dispersion by combining \(U_{\mathrm{dom}}(u)\), the HLC risk map \(R_{\mathrm{HLC}}(u)\), and the normalized dispersion response \(\rho_{\sigma}(u)\):
\begin{equation}
\mathcal{L}_{\mathrm{dom}} =
\frac{
\sum_{u\in\Omega}
R_{\mathrm{HLC}}(u)
U_{\mathrm{dom}}(u)
\rho_{\sigma}(u)
}
{
\sum_{u\in\Omega}
R_{\mathrm{HLC}}(u)+\epsilon
}.
\label{eq:l_dom}
\end{equation}

Here, \(R_{\mathrm{HLC}}(u)\) determines where height-layer optimization should be emphasized, \(U_{\mathrm{dom}}(u)\) identifies dominant hypotheses with insufficient reliability, and \(\rho_{\sigma}(u)\) measures the associated vertical inconsistency. Their joint weighting restricts the correction to locations where unreliable dominance is accompanied by substantial height dispersion. Consequently, \(\mathcal{L}_{\mathrm{dom}}\) does not enforce a globally single-layer surface; instead, it prevents insufficiently reliable dominant evidence from disproportionately guiding Gaussian optimization while largely preserving well-supported height configurations.

\subsection{Secondary-layer Suppression (SLS)}
\label{sec:secondary_layer}

Even after a dominant height hypothesis has been identified, distant secondary layers may retain non-negligible compositing contributions and bias the recovered surface elevation. Such interference is particularly relevant near building boundaries, occlusions, and other height discontinuities, where responses associated with vertically separated surfaces may coexist along the same viewing direction. The secondary-layer suppression (SLS) module therefore regulates non-dominant height hypotheses that are sufficiently separated from the dominant layer and may interfere with elevation estimation.

Given the dominant height hypothesis \(H_{\mathrm{dom}}(u)\), we define the altitude separation and far-layer indicator as
\begin{equation}
\begin{aligned}
\Delta H_k(u)
&=
\left|\bar{h}_k(u)-H_{\mathrm{dom}}(u)\right|,\\
M_{\mathrm{far},k}(u)
&=
\mathbb{I}\left(\Delta H_k(u)>\tau_h\right),
\qquad k\neq m(u),
\end{aligned}
\label{eq:far_layer_indicator}
\end{equation}
where \(\tau_h\) is the same GSD-aware threshold used for candidate-layer construction. \(M_{\mathrm{far},k}(u)\) therefore identifies secondary hypotheses that are geometrically separated from the selected dominant surface.

For each secondary layer, we define a reliability proxy \(Q_{\mathrm{rel},k}(u)=\sqrt{0.5\,Q_{\mathrm{foot},k}(u)}\), where the fixed factor \(0.5\) provides a fixed scaling factor and \(Q_{\mathrm{foot},k}(u)\) reflects the footprint support of the layer. The reliability-weighted secondary response is then
\begin{equation}
B_{\mathrm{sec}}(u)=
\frac{
\sum_{k\neq m(u)}
W_k(u)
M_{\mathrm{far},k}(u)
\left(
1-Q_{\mathrm{rel},k}(u)
\right)
}
{
O(u)+\epsilon
}.
\label{eq:bad_secondary}
\end{equation}
A large \(B_{\mathrm{sec}}(u)\) indicates that distant secondary hypotheses retain substantial compositing contributions despite receiving limited reliability support.

The corresponding secondary-layer suppression loss is
\begin{equation}
\mathcal{L}_{\mathrm{sec}}=
\frac{
\sum_{u\in\Omega}
R_{\mathrm{HLC}}(u)
B_{\mathrm{sec}}(u)
}
{
\sum_{u\in\Omega}
R_{\mathrm{HLC}}(u)+\epsilon
}.
\label{eq:l_sec}
\end{equation}
This term preferentially suppresses weakly supported distant hypotheses in regions emphasized by the HLC risk map.

Reliability-aware suppression alone does not fully control secondary-layer interference, because a distant competing hypothesis may still retain a substantial contribution even when its footprint support is relatively strong. We therefore additionally measure the total contribution of all far secondary layers:
\begin{equation}
F_{\mathrm{far}}(u)=
\frac{
\sum_{k\neq m(u)}
W_k(u)
M_{\mathrm{far},k}(u)
}
{
O(u)+\epsilon
}.
\label{eq:far_layer_ratio}
\end{equation}
The corresponding far-layer loss is defined as
\begin{equation}
\mathcal{L}_{\mathrm{far}}=
\frac{
\sum_{u\in\Omega}
R_{\mathrm{HLC}}(u)
F_{\mathrm{far}}(u)
}
{
\sum_{u\in\Omega}
R_{\mathrm{HLC}}(u)+\epsilon
}.
\label{eq:l_far}
\end{equation}

Although \(\mathcal{L}_{\mathrm{sec}}\) and \(\mathcal{L}_{\mathrm{far}}\) share the same geometric separation criterion, they regulate complementary aspects of secondary-layer interference. \(\mathcal{L}_{\mathrm{sec}}\) emphasizes distant hypotheses with insufficient reliability support, whereas \(\mathcal{L}_{\mathrm{far}}\) limits the overall contribution of distant non-dominant hypotheses regardless of their individual support. Together, these objectives selectively reduce conflicting secondary height evidence and mitigate the intermediate-elevation bias caused by competing height layers.

\begin{algorithm}[t]
\caption{Reliability-aware height-layer optimization in HLC-GS}
\label{alg:hlcgs}
\begin{algorithmic}[1]
\Require Multi-view optical satellite images and imaging models; initial Gaussian representation; total iterations \(T\); HLC activation iteration \(T_{\mathrm{start}}\)
\Ensure Optimized Gaussian representation and reconstructed DSM

\State Initialize the shared Gaussian representation following EOGS

\For{\(t=1\) to \(T\)}
    \State Render the current Gaussian representation from a training view
    \State Compute the EOGS reconstruction loss \(\mathcal{L}_{\mathrm{EOGS}}\)

    \If{\(t \geq T_{\mathrm{start}}\)}
        \State Compute per-pixel height dispersion and construct candidate height layers
        \State Estimate layer competition using \(C_{\mathrm{dom}}(u)\) and \(G_{\mathrm{dom}}(u)\)
        \State Evaluate dominant-hypothesis reliability \(Q_{\mathrm{rel}}(u)\) using layer dominance and footprint support
        \State Construct the HLC risk map \(R_{\mathrm{HLC}}(u)\)
        \State Compute the dominant-layer reliability correction (DRC)
        \State Identify and suppress interfering secondary layers using SLS
        \State Compute the HLC objective \(\mathcal{L}_{\mathrm{HLC}}\) using DRC and SLS
        \State Compute the total loss \(\mathcal{L}=\mathcal{L}_{\mathrm{EOGS}}+\eta(t)\mathcal{L}_{\mathrm{HLC}}\)
    \Else
        \State Set \(\mathcal{L}=\mathcal{L}_{\mathrm{EOGS}}\)
    \EndIf

    \State Update the Gaussian parameters by back-propagation
\EndFor

\State Back-project the rendered surface samples and rasterize them onto the target DSM grid

\end{algorithmic}
\end{algorithm}

\subsection{Overall Objective}
\label{sec:overall_objective}

The proposed height-layer reasoning mechanism is incorporated into the original EOGS objective as a reliability-aware height-layer optimization objective. The complete HLC objective consists of three complementary components:
\begin{equation}
\mathcal{L}_{\mathrm{HLC}} =
\lambda_{\mathrm{dom}}\mathcal{L}_{\mathrm{dom}}
+\lambda_{\mathrm{sec}}\mathcal{L}_{\mathrm{sec}}
+\lambda_{\mathrm{far}}\mathcal{L}_{\mathrm{far}},
\label{eq:l_hlc}
\end{equation}
where \(\mathcal{L}_{\mathrm{dom}}\) corrects unsupported dominant height
hypotheses by penalizing their associated height dispersion,
\(\mathcal{L}_{\mathrm{sec}}\) suppresses weakly supported far secondary
hypotheses, and \(\mathcal{L}_{\mathrm{far}}\) limits the total contribution
of distant non-dominant height layers.
Together, these terms regulate complementary aspects of height-layer
competition: dominant-hypothesis reliability, support-aware secondary-layer
suppression, and residual far-layer contribution control.

In our implementation, the corresponding weights are empirically set to
\(\lambda_{\mathrm{dom}}=0.75\),
\(\lambda_{\mathrm{sec}}=1.25\), and
\(\lambda_{\mathrm{far}}=0.35\), and are kept fixed across all experiments.
The relatively stronger weights assigned to
\(\mathcal{L}_{\mathrm{dom}}\) and
\(\mathcal{L}_{\mathrm{sec}}\) prioritize the correction of unreliable
dominant responses and weakly supported competing secondary hypotheses,
whereas \(\mathcal{L}_{\mathrm{far}}\) provides a weaker global control on
the residual mass of distant height layers.

During the early stage of optimization, Gaussian positions, opacities, and
the resulting height-layer statistics are still unstable.
Applying the HLC constraints too early may therefore interfere with the
formation of the initial scene geometry.
We consequently adopt a delayed activation and linear warm-up strategy:
\begin{equation}
\mathcal{L}=
\mathcal{L}_{\mathrm{EOGS}}
+
\eta(t)\mathcal{L}_{\mathrm{HLC}},
\label{eq:total_loss}
\end{equation}
where \(\mathcal{L}_{\mathrm{EOGS}}\) denotes the original EOGS objective.
The HLC weight \(\eta(t)\) is activated after
\(T_{\mathrm{start}}=3000\) iterations and linearly increased to
\(1.5\times10^{-2}\) over the subsequent \(1000\) iterations.
It then remains fixed at its maximum value for the remainder of training.
This delayed activation allows the shared Gaussian representation to first
establish a coarse geometric structure before reliability-aware height-layer
constraints are introduced, reducing interference with early-stage scene
optimization.

The complete optimization procedure is summarized in
Algorithm~\ref{alg:hlcgs}.

\section{Experiments}
\label{sec:experiments}

\subsection{Datasets and Experimental Design}
\label{subsec:datasets}

We evaluate HLC-GS following the general experimental protocol of EOGS~\citep{aira2025gaussian}. The experiments are organized to assess three complementary aspects of the proposed method. First, quantitative and qualitative comparisons with representative photogrammetric, NeRF-based, Gaussian-based, and surface-aware reconstruction methods, together with local ROI analysis, evaluate the overall quality of the reconstructed surface geometry. Second, component ablation and a series of diagnostic analyses examine the proposed height-layer reasoning mechanism, including the reliability of dominant height hypotheses, the role of training-time height-layer optimization, the distribution of reconstruction improvements across different error levels, and the localization ability of the pre-HLC risk map. Finally, parameter sensitivity and runtime comparisons evaluate the stability and computational efficiency of HLC-GS.

We use seven benchmark scenes from two public satellite datasets: four AOIs from DFC2019~\citep{bosch2019semantic} and three AOIs from IARPA2016~\citep{bosch2016multiple}. Each AOI contains 10--25 cropped, non-orthorectified WorldView-3 images acquired from different viewing geometries and dates, together with the corresponding satellite imaging metadata, including rational polynomial camera (RPC) coefficients and solar-direction information. Following EOGS~\citep{aira2025gaussian}, we use the bundle-adjusted RPC coefficients provided by EO-NeRF~\citep{mari2023multi}. Each scene covers an area of approximately \(256\,\mathrm{m}\times256\,\mathrm{m}\), with a ground sampling distance (GSD) of approximately \(0.5\,\mathrm{m}\) for DFC2019 and \(0.3\,\mathrm{m}\) for IARPA2016.

\subsection{Evaluation Metrics}
\label{subsec:metrics}

Following established satellite surface-reconstruction evaluation protocols~\citep{gao2023general}, we evaluate the reconstructed geometry using Mean Absolute Error (MAE), Root Mean Square Error (RMSE), and Percentage of Accurate Grids (PAG). All metrics are computed over valid non-forest cells of the rasterized DSM. Let \(H_{\mathrm{pred}}(u)\) and \(H_{\mathrm{gt}}(u)\) denote the predicted and reference elevations at an evaluation cell \(u\), respectively. MAE measures the average absolute elevation error, whereas RMSE assigns greater weight to large deviations and is therefore more sensitive to severe reconstruction errors.

PAG measures the proportion of evaluation cells whose absolute elevation error falls within a prescribed tolerance. For an elevation-error threshold \(\tau\), it is defined as
\begin{equation}
\mathrm{PAG}_{\tau}
=
\frac{1}{|\Omega_{\mathrm{eval}}|}
\sum_{u\in\Omega_{\mathrm{eval}}}
\mathbb{I}
\left(
\left|
H_{\mathrm{pred}}(u)-H_{\mathrm{gt}}(u)
\right|
\leq \tau
\right)
\times 100\% ,
\label{eq:pag}
\end{equation}
where \(\Omega_{\mathrm{eval}}\) denotes the set of valid evaluation cells. We report \(\mathrm{PAG}_{2.5}\) and \(\mathrm{PAG}_{7.5}\), corresponding to elevation-error tolerances of \(2.5\,\mathrm{m}\) and \(7.5\,\mathrm{m}\), respectively. While MAE and RMSE characterize the overall magnitude of reconstruction errors, PAG measures the fraction of the reconstructed surface satisfying a specified elevation-accuracy requirement. Accordingly, \(\mathrm{PAG}_{2.5}\) provides a stricter assessment of elevation accuracy, whereas \(\mathrm{PAG}_{7.5}\) reflects robustness to larger reconstruction deviations.

\subsection{Implementation Details}
\label{subsec:implementation}

HLC-GS is implemented in PyTorch based on EOGS~\citep{aira2025gaussian}, with CUDA kernels used for differentiable Gaussian rasterization and view-dependent per-pixel height-layer statistics. For each AOI, Gaussian primitives are initialized following the EOGS~\cite{aira2025gaussian} strategy with a density of \(0.13\) Gaussians per cubic meter and optimized with Adam for \(10{,}000\) iterations. Unless otherwise specified, all EOGS settings, including Gaussian pruning and attribute learning rates, are kept unchanged to isolate the effect of the proposed height-layer optimization. All experiments are conducted on a workstation equipped with an NVIDIA RTX PRO 6000 Blackwell Workstation Edition GPU.

The HLC objective is activated after \(T_{\mathrm{start}}=3000\) iterations, and its global weight \(\eta(t)\) is gradually increased to a maximum value of \(0.015\) over the following \(1000\) iterations. The principal HLC terms are weighted by \(0.75\), \(1.25\), and \(0.35\) for \(\mathcal{L}_{\mathrm{dom}}\), \(\mathcal{L}_{\mathrm{sec}}\), and \(\mathcal{L}_{\mathrm{far}}\), respectively. Multi-temporal satellite imagery often contains substantial seasonal and appearance changes in forest regions, where canopy geometry is also less stable than that of man-made structures. For evaluation, tree-covered regions are excluded using the provided forest masks, following the protocol adopted by EOGS~\citep{aira2025gaussian}, since multi-temporal canopy geometry is less stable and is strongly affected by seasonal appearance variations. All reported MAE, RMSE, PAG$_{2.5}$, and PAG$_{7.5}$ values are therefore computed over valid non-forest DSM cells.

\subsection{Comparative Methods}
\label{subsec:comparative_methods}

We compare HLC-GS with seven representative methods spanning traditional photogrammetry, neural radiance-field reconstruction, and Gaussian-based 3D reconstruction. S2P~\citep{de2014automatic} is included as a classical photogrammetric baseline that combines RPC-based stereo rectification, dense image matching, and surface reconstruction. For neural radiance-field methods, we evaluate SAT-NGP~\citep{billouard2024sat}, Sat-NeRF~\citep{mari2022sat}, and EO-NeRF~\citep{mari2023multi}, representing efficient neural scene modeling, multi-temporal satellite reconstruction, and illumination-aware geometric reconstruction, respectively.

For Gaussian-based reconstruction, we compare against 3DGS~\citep{kerbl20233d}, the surface-oriented 2DGS~\citep{huang20242d}, and the satellite-specific EOGS~\citep{aira2025gaussian}. 3DGS represents the scene using volumetric Gaussian primitives, whereas 2DGS employs oriented Gaussian disks together with depth-distortion and normal-consistency regularization to encourage compact and locally coherent surface geometry. Since the original 3DGS and 2DGS formulations are developed for perspective cameras, their projection modules are adapted to satellite imagery using the same camera approximation framework adopted in our experiments, while preserving their original scene representations and geometric objectives. EOGS is specifically designed for Earth-observation photogrammetry and serves as the primary Gaussian baseline and backbone of HLC-GS.

All methods are evaluated using their released implementations or carefully reproduced implementations under a unified evaluation protocol. Whenever applicable, learning-based methods use the same satellite image subsets and corresponding imaging parameters. The reconstructed surfaces are transformed into a common geospatial reference frame and evaluated using the same spatial extent, valid-region definition, forest and water exclusion masks, rigid registration procedure, and metric computation protocol. This unified evaluation minimizes differences arising from post-processing and enables direct comparison of the recovered surface geometry.

\begin{table*}[!t]
\centering
\caption{Quantitative 3D reconstruction results on the JAX scenes.}
\label{tab:dsm_results_jax}

\footnotesize
\renewcommand{\arraystretch}{1.08}

\setlength{\tabcolsep}{1.6pt}

\begin{tabular*}{\textwidth}{
@{\extracolsep{\fill}}
l*{16}{c}
@{}
}
\toprule

\multirow{2}{*}{Method}
& \multicolumn{4}{c}{JAX 004}
& \multicolumn{4}{c}{JAX 068}
& \multicolumn{4}{c}{JAX 214}
& \multicolumn{4}{c}{JAX 260}
\\

\cmidrule(lr){2-5}
\cmidrule(lr){6-9}
\cmidrule(lr){10-13}
\cmidrule(lr){14-17}

& MAE$\downarrow$
& RMSE$\downarrow$
& PAG$_{2.5}$$\uparrow$
& PAG$_{7.5}$$\uparrow$

& MAE$\downarrow$
& RMSE$\downarrow$
& PAG$_{2.5}$$\uparrow$
& PAG$_{7.5}$$\uparrow$

& MAE$\downarrow$
& RMSE$\downarrow$
& PAG$_{2.5}$$\uparrow$
& PAG$_{7.5}$$\uparrow$

& MAE$\downarrow$
& RMSE$\downarrow$
& PAG$_{2.5}$$\uparrow$
& PAG$_{7.5}$$\uparrow$
\\

\midrule

S2P~\citep{de2014automatic}
& 0.83 & 1.64 & 90.22 & 98.99
& 2.09 & 4.60 & 86.39 & 95.70
& 3.56 & 7.97 & 75.12 & 86.22
& 3.21 & 5.17 & 67.47 & 95.33
\\

SAT-NGP~\citep{billouard2024sat}
& 1.08 & 1.78 & 91.70 & 98.89
& 1.39 & 2.48 & 87.37 & 97.82
& 1.96 & 4.15 & 80.52 & 96.38
& 1.49 & 2.47 & 85.21 & 98.43
\\

3DGS~\citep{kerbl20233d}
& 4.52 & 5.62 & 32.93 & 82.02
& 9.06 & 11.73 & 18.24 & 50.84
& 11.62 & 15.69 & 15.04 & 42.98
& 6.52 & 8.29 & 23.88 & 65.30
\\

2DGS~\citep{huang20242d}
& 0.92 & 1.41 & 93.35 & 99.36
& 2.62 & 4.78 & 77.43 & 93.14
& 5.52 & 8.56 & 44.40 & 77.46
& 2.76 & 4.54 & 69.12 & 91.68
\\

Sat-NeRF~\citep{mari2022sat}
& 4.29 & 5.44 & 37.31 & 82.15
& 2.22 & 3.82 & 76.67 & 95.12
& 3.44 & 6.81 & 62.48 & 90.21
& 4.93 & 6.33 & 31.47 & 79.33
\\

EO-NeRF~\citep{mari2023multi}
& 1.24 & 1.95 & 87.17 & 98.89
& 1.16 & 2.20 & 91.63 & 98.64
& 1.84 & 3.76 & 83.60 & 95.99
& 2.06 & 2.81 & 76.44 & 98.54
\\

EOGS~\citep{aira2025gaussian}
& 0.85 & 1.52 & 91.74 & 99.47
& 1.04 & 2.29 & 90.84 & 98.21
& 1.62 & 3.68 & 85.97 & \textbf{96.21}
& 1.34 & 2.52 & 87.54 & 97.64
\\

HLC-GS
& \textbf{0.67}
& \textbf{1.32}
& \textbf{93.73}
& \textbf{99.48}

& \textbf{0.74}
& \textbf{1.91}
& \textbf{93.62}
& \textbf{98.72}

& \textbf{1.29}
& \textbf{3.57}
& \textbf{88.29}
& 95.52

& \textbf{1.08}
& \textbf{2.17}
& \textbf{89.03}
& \textbf{98.56}
\\

\bottomrule
\end{tabular*}

\end{table*}

\begin{table*}[!t]
\centering
\caption{Quantitative 3D reconstruction results on the IARPA scenes.}
\label{tab:dsm_results_iarpa}

\footnotesize
\renewcommand{\arraystretch}{1.08}

\setlength{\tabcolsep}{2.6pt}

\begin{tabular*}{\textwidth}{
@{\extracolsep{\fill}}
l*{12}{c}
@{}
}
\toprule

\multirow{2}{*}{Method}
& \multicolumn{4}{c}{IARPA 001}
& \multicolumn{4}{c}{IARPA 002}
& \multicolumn{4}{c}{IARPA 003}
\\

\cmidrule(lr){2-5}
\cmidrule(lr){6-9}
\cmidrule(lr){10-13}

& MAE$\downarrow$
& RMSE$\downarrow$
& PAG$_{2.5}$$\uparrow$
& PAG$_{7.5}$$\uparrow$

& MAE$\downarrow$
& RMSE$\downarrow$
& PAG$_{2.5}$$\uparrow$
& PAG$_{7.5}$$\uparrow$

& MAE$\downarrow$
& RMSE$\downarrow$
& PAG$_{2.5}$$\uparrow$
& PAG$_{7.5}$$\uparrow$
\\

\midrule

S2P~\citep{de2014automatic}
& 3.63 & 5.54 & 65.32 & 77.27
& 3.84 & 7.24 & 71.94 & 80.01
& 2.62 & 5.16 & 73.59 & 86.22
\\

SAT-NGP~\citep{billouard2024sat}
& -- & -- & -- & --
& -- & -- & -- & --
& -- & -- & -- & --
\\

3DGS~\citep{kerbl20233d}
& 6.84 & 9.19 & 25.55 & 65.81
& 10.65 & 14.07 & 16.01 & 45.23
& 11.70 & 15.61 & 14.30 & 41.95
\\

2DGS~\citep{huang20242d}
& 2.30 & 3.20 & 63.17 & 98.28
& 3.26 & 4.38 & 57.17 & 89.88
& 2.66 & 4.43 & 66.64 & 90.73
\\

Sat-NeRF~\citep{mari2022sat}
& 2.59 & 3.65 & 63.54 & 93.23
& 4.39 & 6.02 & 42.84 & 81.16
& 4.55 & 8.27 & 58.85 & 81.19
\\

EO-NeRF~\citep{mari2023multi}
& 2.85 & 3.97 & 65.65 & 91.77
& 2.44 & 4.50 & 77.65 & 91.56
& 2.36 & 3.98 & 73.62 & 92.76
\\

EOGS~\citep{aira2025gaussian}
& 1.44 & 2.30 & 82.70 & \textbf{98.42}
& 1.85 & 3.26 & 82.72 & 94.44
& 2.05 & 3.91 & 81.13 & \textbf{94.86}
\\

HLC-GS
& \textbf{1.13}
& \textbf{2.29}
& \textbf{87.41}
& 97.83

& \textbf{1.42}
& \textbf{2.99}
& \textbf{86.90}
& \textbf{95.23}

& \textbf{1.95}
& \textbf{3.82}
& \textbf{81.26}
& 94.46
\\

\bottomrule
\end{tabular*}

\end{table*}

\section{Results and Discussion}
\label{sec:results}

\subsection{Quantitative Results}
\label{subsec:quantitative_results}

Tables~\ref{tab:dsm_results_jax} and~\ref{tab:dsm_results_iarpa} report the quantitative surface-reconstruction results on four JAX scenes and three IARPA scenes, respectively. HLC-GS is compared with representative photogrammetric, NeRF-based, Gaussian-based, and surface-aware reconstruction methods using MAE, RMSE, PAG$_{2.5}$, and PAG$_{7.5}$.

Overall, HLC-GS achieves the best average MAE and RMSE among the compared methods. Compared with the EOGS baseline~\citep{aira2025gaussian}, HLC-GS reduces the average MAE from \(1.46\,\mathrm{m}\) to \(1.18\,\mathrm{m}\) and RMSE from \(2.78\,\mathrm{m}\) to \(2.58\,\mathrm{m}\), corresponding to relative reductions of \(19.18\%\) and \(7.19\%\), respectively. HLC-GS also increases PAG$_{2.5}$ from \(86.09\%\) to \(88.61\%\), an improvement of \(2.52\) percentage points, while maintaining a high PAG$_{7.5}$ of \(97.11\%\). The larger improvement under the stricter \(2.5\,\mathrm{m}\) tolerance indicates that HLC-GS increases the proportion of evaluation cells falling within a more accurate elevation range. Under the more relaxed \(7.5\,\mathrm{m}\) criterion, the seven-scene mean PAG improves slightly, although small decreases are observed in some individual scenes. The surface-oriented 2DGS baseline~\citep{huang20242d} achieves an average MAE of \(2.86\,\mathrm{m}\) and RMSE of \(4.47\,\mathrm{m}\), with PAG$_{2.5}$ and PAG$_{7.5}$ values of \(67.33\%\) and \(91.53\%\), respectively. Its relatively better performance on JAX004 suggests that stronger surface-oriented constraints can be effective when local geometric structures are sufficiently captured. However, its larger performance variation across scenes indicates that improved surface coherence alone does not guarantee stable elevation recovery under the geometric ambiguities present in multi-view satellite imagery. In contrast, HLC-GS maintains consistently strong reconstruction performance across both the JAX and IARPA datasets. The comparison with 2DGS further highlights the distinction between surface organization and height-hypothesis reliability. While 2DGS primarily encourages Gaussian primitives to form a more compact and coherent surface representation, HLC-GS explicitly reasons about the competition, support, and reliability of vertically separated height hypotheses during optimization. These results show that reliability-aware height-layer regulation provides a complementary mechanism for improving Gaussian-based satellite 3D reconstruction beyond generic surface regularization.

\subsection{Qualitative Results}
\label{subsec:qualitative_results}

\begin{figure*}[!t]
\centering
\includegraphics[
  width=\textwidth,
  height=0.7\textheight,
  keepaspectratio
]{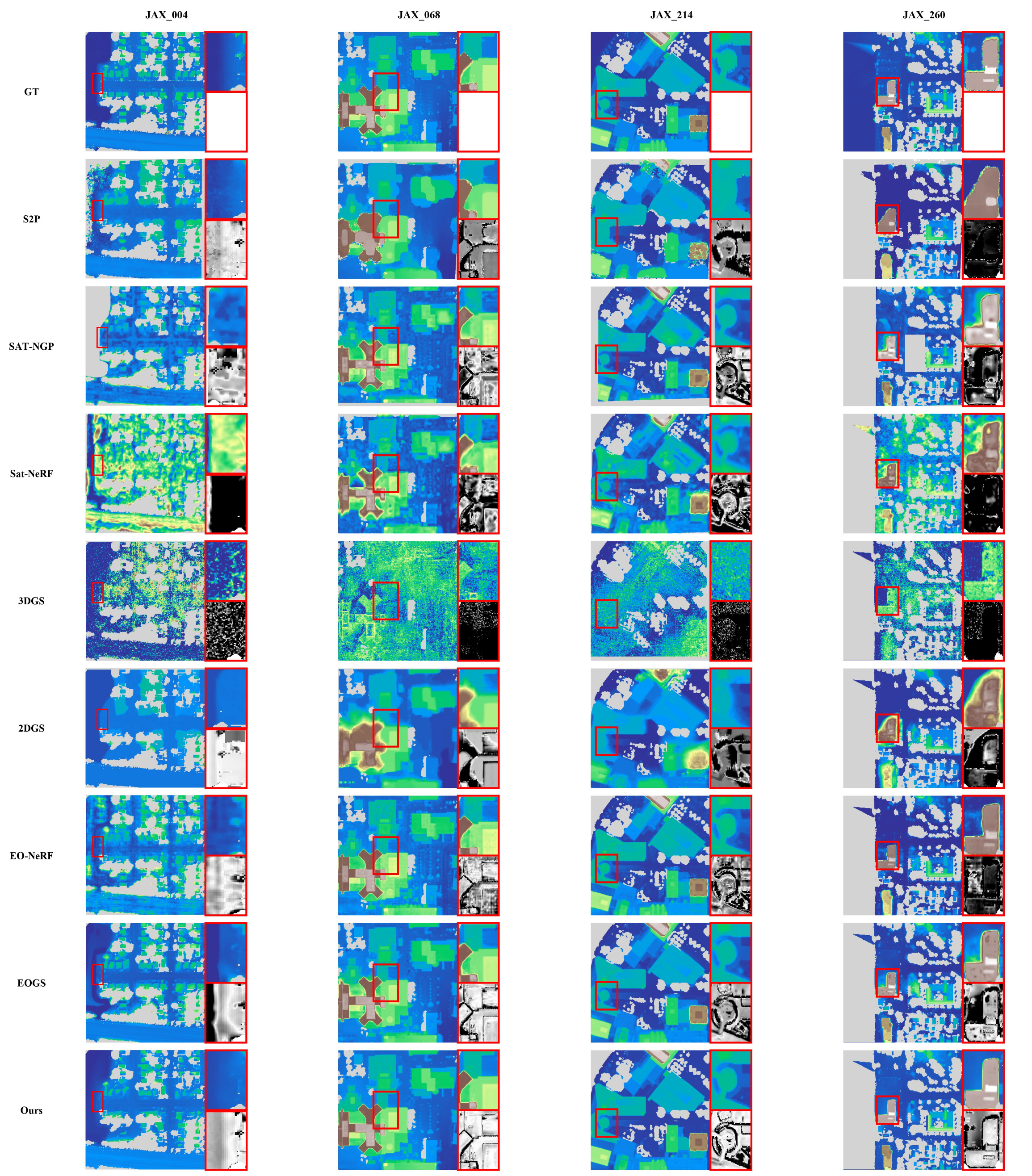}
\caption{Qualitative comparison of 3D reconstruction results on representative JAX scenes. Each group shows the reconstructed DSM, a zoomed-in local region, and the corresponding absolute elevation-error map.}
\label{fig:qualitative_comparison_jax}
\end{figure*}

\begin{figure*}[!t]
\centering
\includegraphics[
  width=\textwidth,
  height=0.75\textheight,
  keepaspectratio
]{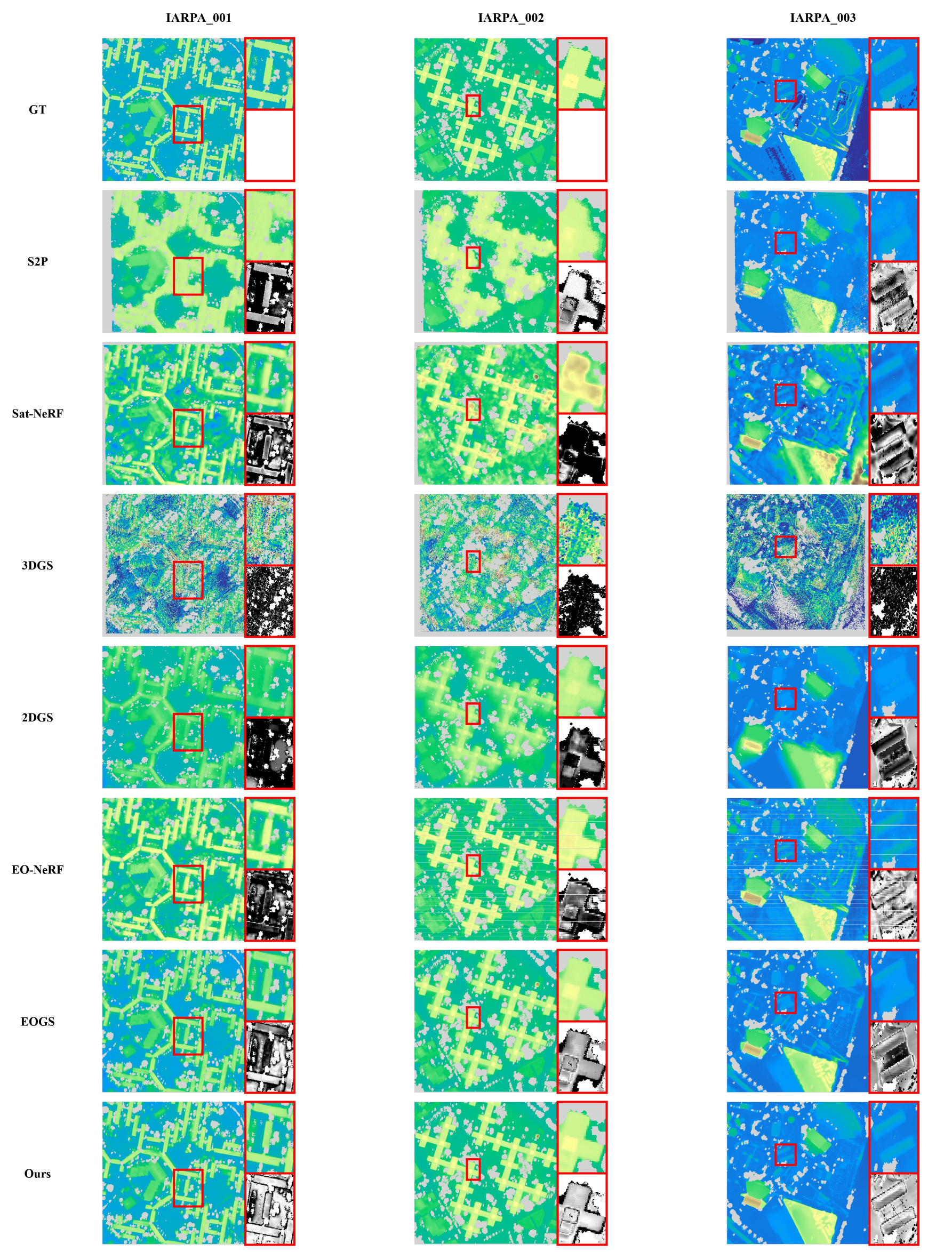}
\caption{Qualitative comparison of 3D reconstruction results on representative IARPA scenes. Each group shows the reconstructed DSM, a zoomed-in local region, and the corresponding absolute elevation-error map.}
\label{fig:qualitative_comparison_iarpa}
\end{figure*}

Figures~\ref{fig:qualitative_comparison_jax} and~\ref{fig:qualitative_comparison_iarpa} present qualitative surface-reconstruction results on the seven evaluated scenes from the JAX and IARPA datasets, respectively. To facilitate detailed comparison, a representative local region is selected for each scene, and the corresponding reconstructed DSM, zoomed-in surface geometry, and absolute elevation-error map are visualized. In the error maps, lighter intensities indicate smaller absolute elevation errors, whereas darker intensities indicate larger errors. The same visualization range is used for all methods within each scene to ensure consistent comparison.

The local reconstructions reveal that several comparative methods retain noticeable geometric errors near building boundaries and roof--ground transitions. These errors are often manifested as blurred roof structures, distorted building boundaries, or mixed elevations caused by vertically separated height responses, and are reflected by stronger responses in the corresponding elevation-error maps.

In contrast, HLC-GS generally reconstructs sharper and more complete building structures while reducing elevation errors in the selected regions. Improvements are particularly visible around roof boundaries and roof--ground transitions, where competing height responses are more likely to affect surface recovery. The corresponding error maps exhibit fewer high-error regions, consistent with the quantitative improvements reported in Tables~\ref{tab:dsm_results_jax} and~\ref{tab:dsm_results_iarpa}. These qualitative results further support the effectiveness of reliability-aware height-layer reasoning in improving the geometric quality of Gaussian-based satellite 3D reconstruction.

\begin{table}[t]
\centering
\caption{Local ROI comparison between EOGS and HLC-GS on representative building-boundary regions.}
\label{tab:local_roi}

\footnotesize
\renewcommand{\arraystretch}{1.08}
\setlength{\tabcolsep}{2.0pt}

\begin{tabular*}{0.95\columnwidth}{
@{\extracolsep{\fill}}
llccccc
@{}
}
\toprule

Scene
& Method
& MAE$\downarrow$
& RMSE$\downarrow$
& PAG$_{1}$$\uparrow$
& PAG$_{2.5}$$\uparrow$
& PAG$_{7.5}$$\uparrow$
\\

\midrule

\multirow{2}{*}{JAX\_068}
& EOGS~\citep{aira2025gaussian}
& 2.15
& 4.15
& 61.49
& 73.87
& 92.32
\\

& HLC-GS
& \textbf{1.49}
& \textbf{3.32}
& \textbf{71.89}
& \textbf{86.00}
& \textbf{94.75}
\\

\midrule

\multirow{2}{*}{IARPA\_002}
& EOGS~\citep{aira2025gaussian}
& 2.31
& 3.87
& 48.83
& 78.08
& 90.96
\\

& HLC-GS
& \textbf{1.60}
& \textbf{3.72}
& \textbf{77.98}
& \textbf{86.07}
& \textbf{92.72}
\\

\bottomrule
\end{tabular*}

\end{table}

\subsection{Local ROI Error Analysis}
\label{sec:local_roi}

\begin{figure}[t]
\centering

\begin{subfigure}{0.8\linewidth}
    \centering
    \includegraphics[width=\linewidth]{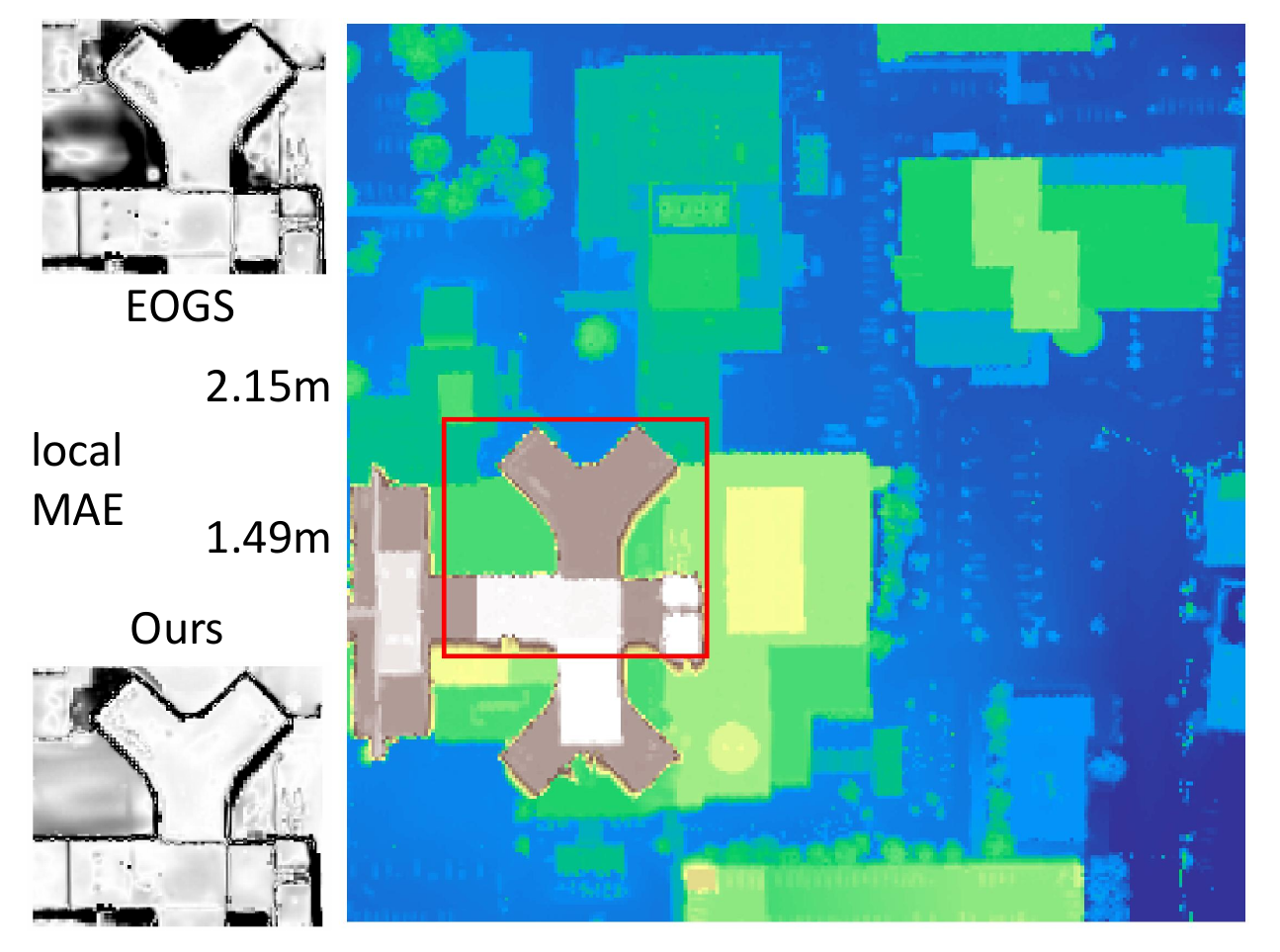}
    \caption{JAX\_068}
    \label{fig:local_roi_jax068}
\end{subfigure}

\vspace{0.3em}

\begin{subfigure}{0.8\linewidth}
    \centering
    \includegraphics[width=\linewidth]{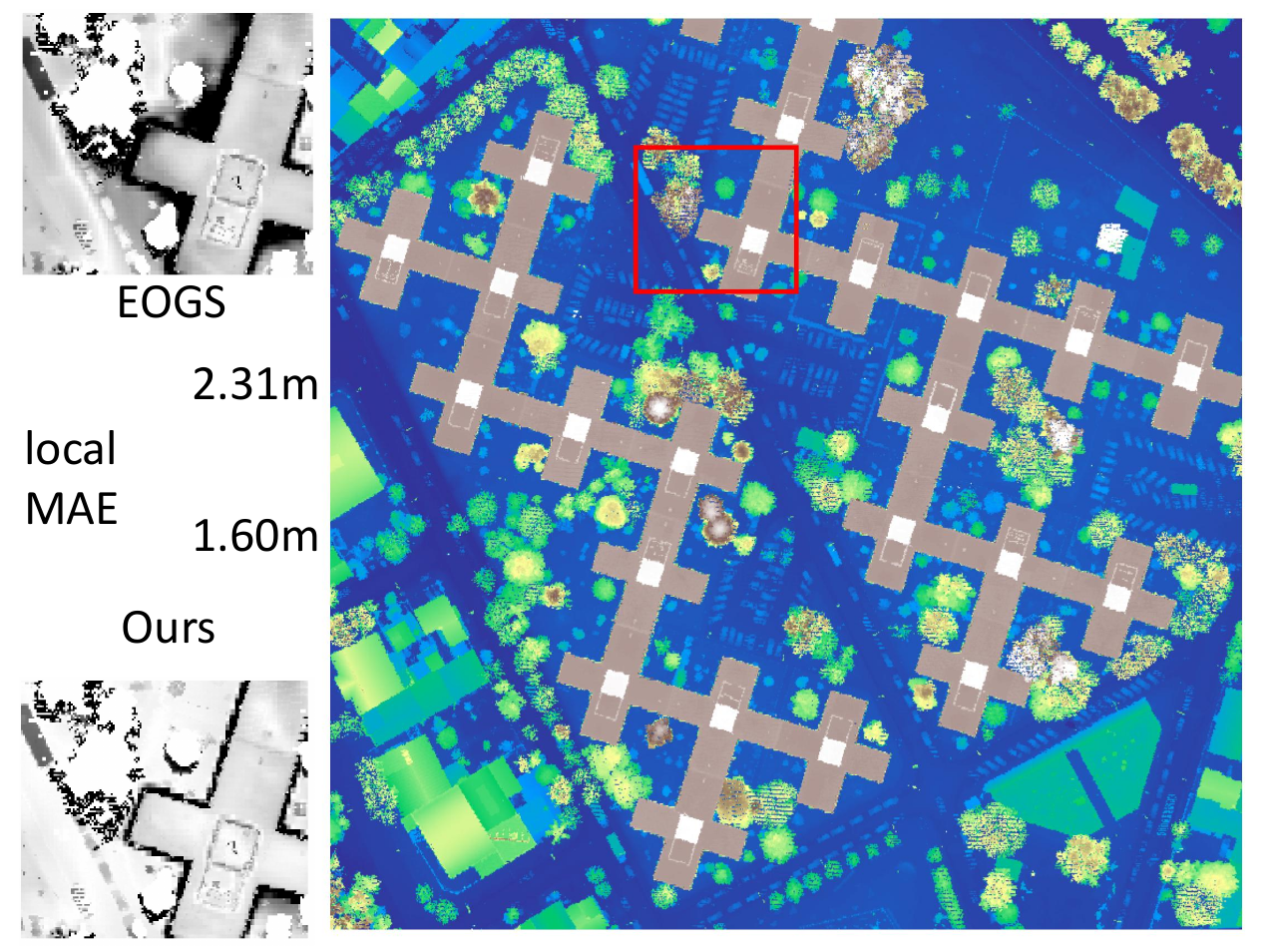}
    \caption{IARPA\_002}
    \label{fig:local_roi_iarpa002}
\end{subfigure}

\caption{
Local ROI comparison in representative building-boundary regions.
Absolute elevation errors of EOGS~\citep{aira2025gaussian} and HLC-GS are visualized using a common error range, with lighter intensities indicating smaller errors.
}
\label{fig:local_roi_comparison}
\end{figure}

To further examine the localized geometric improvements observed in the qualitative results, we conduct an ROI-based error analysis on representative building-boundary regions. These regions contain pronounced roof--ground transitions and other height discontinuities, where competing responses from vertically separated surfaces are more likely to affect elevation recovery. EOGS and HLC-GS are evaluated over exactly the same DSM grid cells using the same evaluation protocol as in the main experiments.

As shown in Fig.~\ref{fig:local_roi_comparison}, HLC-GS exhibits visibly smaller elevation errors than EOGS~\citep{aira2025gaussian} around building boundaries and roof--ground transitions. The reduction of concentrated errors in these regions is consistent with the intended role of height-layer reasoning in limiting the influence of competing height responses from vertically separated layers during surface reconstruction.

The corresponding quantitative results are reported in Table~\ref{tab:local_roi}. For the selected JAX ROI, HLC-GS reduces the MAE from \(2.15\,\mathrm{m}\) to \(1.49\,\mathrm{m}\), corresponding to a relative reduction of \(30.70\%\). For the IARPA ROI, the MAE decreases from \(2.31\,\mathrm{m}\) to \(1.60\,\mathrm{m}\), corresponding to a reduction of \(30.74\%\). These results complement the scene-level evaluation by showing that the improvements of HLC-GS are also evident in representative height-discontinuous regions where elevation ambiguity is particularly pronounced.

\subsection{Ablation Study}
\label{subsec:ablation}

The component removal study further reveals the functional roles of RM, DRC, and SLS. The quantitative results are reported in Table~\ref{tab:ablation}.

\begin{itemize}
    \item Variant (a) corresponds to the EOGS baseline~\citep{aira2025gaussian}. 
    This variant yields the lowest overall accuracy, with an MAE of 1.46 m and an RMSE of 2.78 m. 
    Compared with the HLC-GS model, its MAE increases by 0.28 m and PAG$_{2.5}$ decreases by 2.52 percentage points. 
    This suggests that alpha-weighted Gaussian height aggregation alone may be insufficient to ensure geometric consistency in DSM reconstruction, motivating the need for additional geometric regularization.

    \item Variant (b) removes RM while retaining DRC and SLS. Its MAE increases from 1.18 m to $1.29\,\mathrm{m}$, and $\mathrm{PAG}_{2.5}$ decreases from $88.61\%$ to $88.05\%$. This result indicates that the risk map helps localize high-risk pixels and improves the effectiveness of the height-layer constraints.

    \item Variant (c) removes DRC while retaining RM and SLS. This variant obtains an MAE of \(1.25\,\mathrm{m}\), but its RMSE increases to \(2.68\,\mathrm{m}\) and PAG$_{7.5}$ decreases to \(96.91\%\). These results indicate that DRC mainly contributes to reducing large elevation deviations by regularizing unreliable dominant-layer responses during training.

    \item Variant (d) removes SLS while retaining RM and DRC. 
    Its MAE increases to 1.43 m and PAG$_{2.5}$ decreases to 86.70\%, indicating that weakly supported secondary-layer responses are an important source of DSM errors. This supports the importance of suppressing secondary-layer responses that are inconsistent with the dominant surface for alleviating height-layer mixing.

    \item Variant (e) integrates all three modules and achieves the best overall performance, with an MAE of 1.18 m, an RMSE of 2.58 m, PAG$_{2.5}$ of 88.61\%, and PAG$_{7.5}$ of 97.11\%. 
    These results demonstrate that RM, DRC, and SLS are complementary and jointly improve DSM height accuracy.
\end{itemize}

In summary, RM localizes pixels with high height-layer mixing risk, DRC regularizes unreliable dominant-layer responses, and SLS suppresses inconsistent secondary height-layer responses.
Together, these modules alleviate the height-layer mixing problem caused by alpha-weighted aggregation of Gaussian elevations, thereby improving the geometric accuracy of satellite image-based DSM reconstruction.

\begin{table*}[t]
\centering
\caption{
Ablation study of the proposed modules. Red values indicate performance degradation relative to the full HLC-GS model.
}
\label{tab:ablation}

\footnotesize
\renewcommand{\arraystretch}{1.08}
\setlength{\tabcolsep}{6pt}

\begin{tabular}{c|ccc|cccc}
\toprule

\multirow{2}{*}{Variants}
& \multicolumn{3}{c|}{Components}
& \multirow{2}{*}{MAE $\downarrow$}
& \multirow{2}{*}{RMSE $\downarrow$}
& \multirow{2}{*}{PAG$_{2.5}$ $\uparrow$}
& \multirow{2}{*}{PAG$_{7.5}$ $\uparrow$}
\\

\cline{2-4}

& RM & DRC & SLS
& & & &
\\

\midrule

(a) EOGS
& $\times$
& $\times$
& $\times$
& 1.46{\color{red}$\uparrow$0.28}
& 2.78{\color{red}$\uparrow$0.20}
& 86.09{\color{red}$\downarrow$2.52}
& 97.03{\color{red}$\downarrow$0.08}
\\

(b) w/o RM
& $\times$
& $\checkmark$
& $\checkmark$
& 1.29{\color{red}$\uparrow$0.11}
& 2.63{\color{red}$\uparrow$0.05}
& 88.05{\color{red}$\downarrow$0.56}
& 97.08{\color{red}$\downarrow$0.03}
\\

(c) w/o DRC
& $\checkmark$
& $\times$
& $\checkmark$
& 1.25{\color{red}$\uparrow$0.07}
& 2.68{\color{red}$\uparrow$0.10}
& 88.34{\color{red}$\downarrow$0.27}
& 96.91{\color{red}$\downarrow$0.20}
\\

(d) w/o SLS
& $\checkmark$
& $\checkmark$
& $\times$
& 1.43{\color{red}$\uparrow$0.25}
& 2.70{\color{red}$\uparrow$0.12}
& 86.70{\color{red}$\downarrow$1.91}
& 97.01{\color{red}$\downarrow$0.10}
\\

(e) HLC-GS
& $\checkmark$
& $\checkmark$
& $\checkmark$
& \textbf{1.18}
& \textbf{2.58}
& \textbf{88.61}
& \textbf{97.11}
\\

\bottomrule
\end{tabular}

\end{table*}

\subsection{Analysis of Height-Layer Consistency}

\subsubsection{Reliability of Dominant Height Hypotheses}
\label{subsubsec:dominant_reliability}

A central motivation of DRC is that winning the local height-layer competition does not necessarily indicate that the selected hypothesis corresponds to a reliable surface elevation. We therefore investigate how the dominant-layer confidence \(C_{\mathrm{dom}}\) relates to the accuracy of the selected dominant height. For each valid DSM grid cell, the dominant-height hypothesis is grouped into five confidence intervals according to \(C_{\mathrm{dom}}\), and its elevation error is evaluated against the reference DSM. For comparison, the EOGS elevation error is computed over exactly the same set of grid cells within each interval.

\begin{table}[t]
\centering
\caption{
Accuracy of dominant-height hypotheses across different confidence intervals, with EOGS evaluated on the same cells for reference.
}
\label{tab:confidence_accuracy}

\footnotesize
\renewcommand{\arraystretch}{1.08}
\setlength{\tabcolsep}{2.5pt}

\begin{tabular*}{0.95\columnwidth}{
@{\extracolsep{\fill}}
cccc
@{}
}
\toprule

Confidence
& Coverage (\%)
& EOGS MAE (m)
& Dominant MAE (m)
\\

\midrule

0.0--0.2 & 0.51  & 6.03 & 6.91 \\
0.2--0.4 & 12.86 & 2.92 & 3.38 \\
0.4--0.6 & 20.18 & 1.90 & 1.96 \\
0.6--0.8 & 19.91 & 1.51 & 1.47 \\
0.8--1.0 & 46.54 & 0.86 & 0.82 \\

\bottomrule
\end{tabular*}

\end{table}

As shown in Table~\ref{tab:confidence_accuracy}, the accuracy of the selected dominant height improves consistently as \(C_{\mathrm{dom}}\) increases. In the low-confidence intervals, directly adopting the dominant hypothesis can be less accurate than the original EOGS elevation. For example, in the \(0.2\)--\(0.4\) interval, the dominant-height MAE is \(3.38\,\mathrm{m}\), compared with \(2.92\,\mathrm{m}\) for EOGS over the same cells. As the confidence increases, the dominant-height MAE decreases monotonically, reaching \(0.82\,\mathrm{m}\) in the \(0.8\)--\(1.0\) interval. In the two highest-confidence intervals, the selected dominant heights become slightly more accurate than the corresponding EOGS estimates.

These results demonstrate that dominant-layer confidence provides useful information about the quality of the selected height hypothesis, but local dominance alone is not sufficient to establish reliability. A hypothesis may win the local competition while still yielding an inaccurate surface elevation, particularly when its support is weak or conflicting geometric evidence remains. This observation motivates the reliability-aware design of DRC, where dominant-layer competition is combined with complementary geometric and footprint-support cues before an unreliable dominant response is regulated during optimization.

\subsubsection{Role of Training-time Height-Layer Optimization}
\label{subsubsec:replacement}

We further investigate whether the improvement of HLC-GS can be reproduced by modifying only the final elevation readout, without incorporating height-layer reasoning into Gaussian optimization. Starting from a converged EOGS model, we replace the original alpha-composited elevation \(H^{\mathrm{EOGS}}(u)\) with the extracted dominant-layer elevation \(H^{\mathrm{dom}}(u)\) only when the dominant hypothesis satisfies the prescribed confidence and layer-separation criteria. The resulting elevation estimates are then converted into DSMs using the same surface-reconstruction procedure, rigid registration, and strict common-mask evaluation protocol as the original EOGS output.

\begin{table}[t]
\centering
\caption{
Comparison of post-hoc dominant-height replacement and training-time HLC optimization under the strict common-mask protocol.
}
\label{tab:replacement}

\footnotesize
\renewcommand{\arraystretch}{1.08}
\setlength{\tabcolsep}{2.2pt}

\begin{tabular*}{0.95\columnwidth}{
@{\extracolsep{\fill}}
lcccc
@{}
}
\toprule

Method
& MAE (m)
& RMSE (m)
& PAG$_{2.5}$ (\%)
& PAG$_{7.5}$ (\%)
\\

\midrule

EOGS
& 1.46
& 2.74
& 86.25
& 97.07
\\

EOGS + Replacement
& 1.44
& 2.80
& 86.70
& 96.90
\\

HLC-GS
& \textbf{1.18}
& \textbf{2.58}
& \textbf{88.61}
& \textbf{97.11}
\\

\bottomrule
\end{tabular*}

\end{table}

As shown in Table~\ref{tab:replacement}, post-hoc replacement yields only a marginal reduction in MAE, from \(1.46\,\mathrm{m}\) to \(1.44\,\mathrm{m}\), while RMSE increases from \(2.74\,\mathrm{m}\) to \(2.80\,\mathrm{m}\) and PAG$_{7.5}$ decreases from \(97.07\%\) to \(96.90\%\). In contrast, HLC-GS reduces the MAE to \(1.18\,\mathrm{m}\) and RMSE to \(2.58\,\mathrm{m}\), while simultaneously improving both PAG metrics.

These results indicate that the improvement of HLC-GS cannot be attributed to selecting a different elevation hypothesis only at the final reconstruction stage. Post-hoc replacement changes the surface-elevation readout but leaves the learned Gaussian representation unchanged, including conflicting responses that are not explicitly regulated during optimization. HLC-GS instead incorporates height-layer competition and reliability reasoning into training, allowing unreliable dominant responses and interfering secondary responses to influence the optimization of the Gaussian representation itself. The substantially greater improvement of HLC-GS therefore supports the importance of training-time height-layer regulation rather than post-hoc elevation selection alone.

\subsubsection{Error-Stratified Analysis}
\label{subsubsec:error_stratified}

We further investigate how the effect of HLC-GS varies with the reconstruction difficulty of different surface regions. For each scene, valid DSM cells are ranked according to the absolute elevation error of EOGS and divided into four equally sized quartiles, ranging from the lowest-error group Q1 to the highest-error group Q4. This stratification is used only for retrospective analysis and does not participate in HLC-GS training or inference.

\begin{table}[t]
\centering
\caption{
Error-stratified comparison between EOGS and HLC-GS. Q1--Q4 denote EOGS error quartiles from the lowest to the highest baseline error, and positive $\Delta$MAE indicates improvement over EOGS.
}
\label{tab:error_quartile}

\footnotesize
\renewcommand{\arraystretch}{1.08}
\setlength{\tabcolsep}{2.8pt}

\begin{tabular*}{0.95\columnwidth}{
@{\extracolsep{\fill}}
cccc
@{}
}
\toprule

Quartile
& EOGS MAE (m)
& HLC-GS MAE (m)
& $\Delta$MAE (m)
\\

\midrule

Q1 & 0.25 & 0.45 & -0.20 \\
Q2 & 0.59 & 0.39 &  0.20 \\
Q3 & 0.95 & 0.63 &  0.32 \\
Q4 & 4.05 & 3.20 &  0.85 \\

\bottomrule
\end{tabular*}

\end{table}

As shown in Table~\ref{tab:error_quartile}, the reconstruction gains of HLC-GS become substantially larger in regions with higher baseline errors. HLC-GS reduces the MAE by \(0.20\,\mathrm{m}\), \(0.32\,\mathrm{m}\), and \(0.85\,\mathrm{m}\) in Q2, Q3, and Q4, respectively, with the largest improvement observed in the highest-error quartile. In contrast, the MAE in Q1 increases from \(0.25\,\mathrm{m}\) to \(0.45\,\mathrm{m}\). This result indicates that the proposed height-layer regulation is most beneficial in moderately and severely erroneous regions, while unnecessary intervention may occur in locations that are already reconstructed accurately.

\begin{figure}[t]
\centering
\includegraphics[width=\columnwidth]{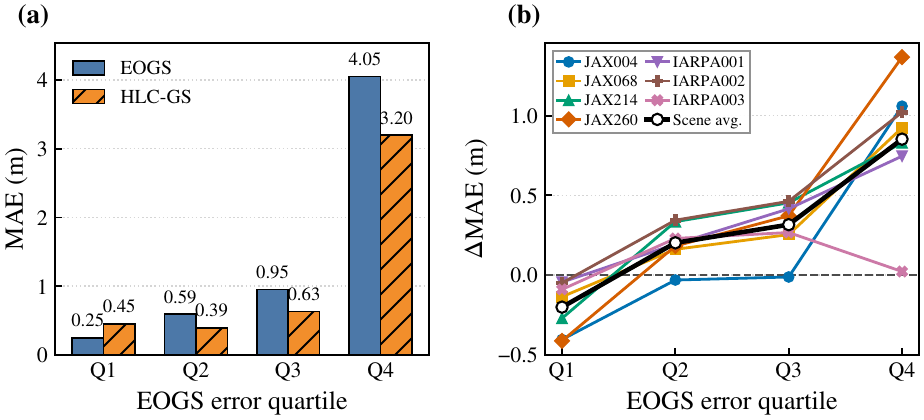}
\caption{
Error-stratified performance of HLC-GS.
(a) MAE across EOGS error quartiles.
(b) Scene-wise MAE improvement over EOGS, with positive values indicating gains.
}
\label{fig:error_quartile_analysis}
\end{figure}

Fig.~\ref{fig:error_quartile_analysis}(b) further shows that this behavior is consistent across all seven evaluated scenes: HLC-GS improves Q2--Q4 in every case, whereas Q1 consistently exhibits a moderate degradation. These results suggest that HLC-GS primarily corrects regions with moderate to large reconstruction errors rather than uniformly modifying the entire surface. At the same time, the degradation observed in Q1 reveals that the current reliability mechanism does not completely avoid over-correction in already accurate regions, leaving room for more selective reliability estimation in future work.

\subsubsection{Pre-HLC Risk Localization Analysis}
\label{subsubsec:risk_localization}

We further investigate whether the proposed risk map can identify regions that subsequently benefit more from HLC optimization before the regularization is applied. To avoid the influence from HLC-induced changes, \(R_{\mathrm{HLC}}\) is computed from the unregularized EOGS representation immediately before HLC activation. For each scene, valid DSM cells are ranked according to their pre-HLC risk values, and the top 10\% and top 20\% regions are used to characterize the high-risk portion of the reconstructed surface. All comparisons follow the same strict common-mask protocol used for the EOGS/HLC-GS evaluation.

\begin{figure}[t]
\centering
\includegraphics[width=\columnwidth]{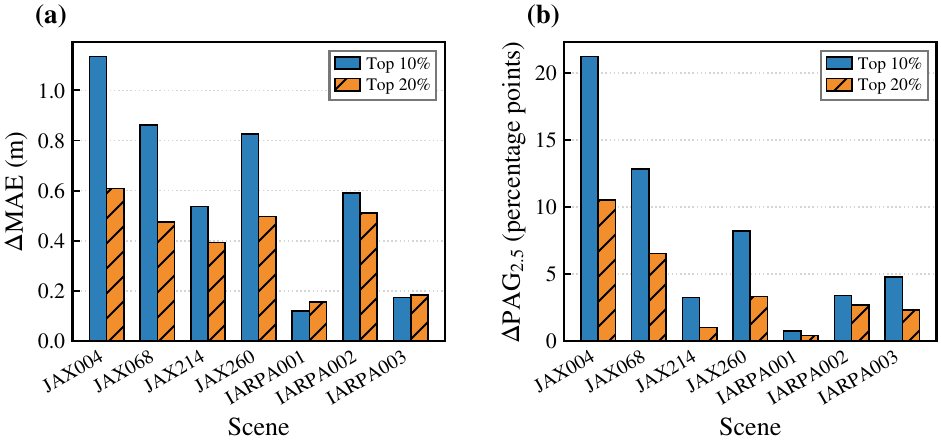}
\caption{Pre-HLC risk localization analysis. (a) Scene-wise MAE reduction and (b) PAG$_{2.5}$ improvement within the top 10\% and top 20\% pre-HLC risk regions. Positive values indicate improvement of HLC-GS over EOGS.}
\label{fig:prehlc_risk_analysis}
\end{figure}

As shown in Fig.~\ref{fig:prehlc_risk_analysis}, HLC-GS improves the high-risk regions across all seven evaluated scenes, with the top 10\% risk region generally exhibiting larger gains than the broader top 20\% region. Within the top 10\% pre-HLC risk region, the MAE decreases from \(1.94\,\mathrm{m}\) for EOGS to \(1.33\,\mathrm{m}\) for HLC-GS, corresponding to a reduction of \(0.61\,\mathrm{m}\). In comparison, the complementary 90\% region improves from \(1.41\,\mathrm{m}\) to \(1.15\,\mathrm{m}\), yielding a smaller reduction of \(0.26\,\mathrm{m}\). Accordingly, the MAE reduction in the highest-risk 10\% region is approximately \(2.35\times\) that observed in its complement.

These results indicate that the pre-HLC risk map preferentially identifies regions in which subsequent height-layer optimization produces larger reconstruction gains. Importantly, this localization is obtained before HLC regularization is introduced and without using reference elevation errors to define the high-risk regions. The risk map therefore provides an effective spatial prioritization signal, assigning greater optimization emphasis to ambiguous and unreliable height configurations rather than imposing the same geometric regularization uniformly across the reconstructed surface.

\subsubsection{Comparison with Height-Dispersion Regularization}
\label{subsubsec:dispersion_regularization}

To examine whether the improvement of HLC-GS can be explained by directly suppressing vertical dispersion, we construct a same-backbone baseline that regularizes the weighted height dispersion while keeping the EOGS representation, input imagery, training schedule, and evaluation protocol unchanged. This comparison isolates the effect of generic height-dispersion regularization from the reliability-aware height-hypothesis regulation introduced in HLC-GS.

\begin{table}[t]
\centering
\caption{
Comparison with height-dispersion regularization under the same EOGS backbone and evaluation protocol.
}
\label{tab:dispersion_regularization}

\footnotesize
\renewcommand{\arraystretch}{1.08}
\setlength{\tabcolsep}{2.8pt}

\begin{tabular*}{0.95\columnwidth}{
@{\extracolsep{\fill}}
lccc
@{}
}
\toprule

Method
& MAE (m) $\downarrow$
& RMSE (m) $\downarrow$
& PAG@2.5 (\%) $\uparrow$
\\

\midrule

EOGS & 1.46 & 2.78 & 86.09 \\
Height-dispersion reg. & 1.58 & 3.17 & 86.99 \\
HLC-GS & \textbf{1.18} & \textbf{2.58} & \textbf{88.61} \\

\bottomrule
\end{tabular*}

\end{table}

As shown in Table~\ref{tab:dispersion_regularization}, directly regularizing height dispersion does not reproduce the overall accuracy improvement achieved by HLC-GS. The height-dispersion baseline obtains an average MAE of 1.58~m and RMSE of 3.17~m, compared with 1.46~m and 2.78~m for EOGS, whereas HLC-GS reduces them to 1.18~m and 2.58~m, respectively. HLC-GS also increases PAG@2.5 from 86.09\% for EOGS to 88.61\%. These results indicate that directly penalizing vertical dispersion alone is insufficient to account for the improvement of HLC-GS. Instead of uniformly suppressing vertically dispersed responses, HLC-GS selectively regulates competing height hypotheses according to their relative support and reliability.

\subsection{Efficiency Comparison}
\label{subsec:efficiency}

\begin{table*}[t]
\centering
\caption{Runtime comparison of different satellite surface-reconstruction methods.}
\label{tab:runtime}

\footnotesize
\renewcommand{\arraystretch}{1.08}
\setlength{\tabcolsep}{3.5pt}

\begin{tabular*}{0.90\textwidth}{
@{\extracolsep{\fill}}
l*{8}{c}
@{}
}
\toprule

Method
& S2P~\citep{de2014automatic}
& SAT-NGP~\citep{billouard2024sat}
& Sat-NeRF~\citep{mari2022sat}
& EO-NeRF~\citep{mari2023multi}
& 3DGS~\citep{kerbl20233d}
& 2DGS~\citep{huang20242d}
& EOGS~\citep{aira2025gaussian}
& HLC-GS
\\

\midrule

Runtime (min)
& 7.52
& 23.13
& 56.03
& 748.80
& 1.43
& 11.40
& 7.81
& 7.22
\\

\bottomrule
\end{tabular*}

\end{table*}

Table~\ref{tab:runtime} compares the average scene-level runtime of
HLC-GS with the evaluated reconstruction methods. For most methods,
the reported runtime is averaged over all seven AOIs, whereas the
SAT-NGP runtime is averaged over the four JAX scenes for which it is
evaluated.

The compared methods exhibit substantial differences in computational cost. The NeRF-based methods generally require longer processing times, with SAT-NGP~\citep{billouard2024sat}, Sat-NeRF~\citep{mari2022sat}, and EO-NeRF~\citep{mari2023multi} requiring \(23.13\), \(56.03\), and \(748.80\) minutes per scene, respectively. Among the Gaussian-based methods, 3DGS~\citep{kerbl20233d} is the fastest at \(1.43\) minutes per scene, while 2DGS~\citep{huang20242d} and EOGS~\citep{aira2025gaussian} require \(11.40\) and \(7.81\) minutes, respectively.

HLC-GS requires \(7.22\) minutes per scene on average, placing it in essentially the same runtime range as EOGS despite the additional computation required for height-layer statistics, reliability estimation, and height-layer regularization. Its runtime is also comparable to that of the traditional S2P pipeline, which requires \(7.52\) minutes per scene. These results indicate that incorporating reliability-aware height-layer reasoning does not introduce substantial computational overhead in practice and largely preserves the computational efficiency of the underlying Gaussian-based reconstruction framework.

\subsection{Parameter Sensitivity Analysis}
\label{subsec:param_sensitivity}

We evaluate the sensitivity of HLC-GS to the key dominant-layer decision thresholds and the maximum global HLC weight on JAX\_068. All variants follow the same training and evaluation protocol as the main experiments. For each experiment, one parameter is varied while the remaining parameters are fixed at their default values.

We first examine the dominant-layer confidence threshold \(\tau_c\), the layer-gap threshold \(\tau_g\), and the reliability threshold \(\tau_u\). Table~\ref{tab:threshold_sensitivity} summarizes the resulting MAE ranges.

\begin{table}[t]
\centering
\caption{
Sensitivity of HLC-GS to dominant-layer decision thresholds on JAX\_068, with each parameter varied independently.
}
\label{tab:threshold_sensitivity}

\footnotesize
\renewcommand{\arraystretch}{1.08}
\setlength{\tabcolsep}{3.5pt}

\begin{tabular*}{0.95\columnwidth}{
@{\extracolsep{\fill}}
ccc
@{}
}
\toprule

Parameter
& Tested values
& MAE range (m)
\\

\midrule

\(\tau_c\)
& 0.35, 0.40, 0.45, 0.50, 0.55
& 0.74--0.80
\\

\(\tau_g\)
& 0.05, 0.10, 0.15, 0.20, 0.25
& 0.74--0.78
\\

\(\tau_u\)
& 0.45, 0.50, 0.55, 0.60, 0.65
& 0.74--0.78
\\

\bottomrule
\end{tabular*}

\end{table}

Across the evaluated threshold ranges, the MAE remains between \(0.74\,\mathrm{m}\) and \(0.80\,\mathrm{m}\). For reference, EOGS obtains an MAE of \(1.04\,\mathrm{m}\) on the same scene. All tested threshold settings therefore preserve a clear improvement over the EOGS baseline, while the relatively small variation in MAE indicates that the reconstruction performance remains reasonably stable around the default parameter settings rather than relying on a narrowly tuned threshold value.

We next vary the maximum global HLC weight while keeping the remaining parameters unchanged. The results are reported in Table~\ref{tab:weight_sensitivity}.

\begin{table}[t]
\centering
\caption{
Sensitivity of HLC-GS to the maximum global HLC loss weight on JAX\_068.
}
\label{tab:weight_sensitivity}

\footnotesize
\renewcommand{\arraystretch}{1.08}
\setlength{\tabcolsep}{2.5pt}

\begin{tabular*}{0.95\columnwidth}{
@{\extracolsep{\fill}}
ccccc
@{}
}
\toprule

HLC weight
& MAE (m)
& RMSE (m)
& PAG$_{2.5}$ (\%)
& PAG$_{7.5}$ (\%)
\\

\midrule

$1.0{\times}10^{-2}$
& 0.77
& 2.02
& 93.53
& 98.67
\\

$1.5{\times}10^{-2}$
& \textbf{0.74}
& \textbf{1.91}
& \textbf{93.62}
& \textbf{98.72}
\\

$1.875{\times}10^{-2}$
& 0.76
& 1.93
& 93.53
& 98.69
\\

$2.25{\times}10^{-2}$
& 0.75
& 1.92
& 93.47
& 98.68
\\

$3.0{\times}10^{-2}$
& 0.76
& 1.95
& 93.46
& 98.61
\\

\bottomrule
\end{tabular*}

\end{table}

As shown in Table~\ref{tab:weight_sensitivity}, the reconstruction accuracy also remains relatively stable across the evaluated HLC-weight range. The MAE varies from \(0.74\,\mathrm{m}\) to \(0.77\,\mathrm{m}\), the RMSE from \(1.91\,\mathrm{m}\) to \(2.02\,\mathrm{m}\), and PAG$_{2.5}$ remains close to \(93.5\%\). The default maximum weight of \(1.5\times10^{-2}\) achieves the best result in this experiment, while nearby settings yield comparable reconstruction accuracy.

Overall, the results on JAX\_068 suggest that HLC-GS is reasonably stable within the evaluated ranges of the dominant-layer thresholds and global HLC weight. We use \(\tau_c=0.45\), \(\tau_g=0.15\), \(\tau_u=0.55\), and a maximum HLC weight of \(1.5\times10^{-2}\) for all main experiments.

\subsection{Discussion}
\label{subsec:discussion}

The experiments suggest that the benefit of HLC-GS arises from selectively regulating ambiguous height configurations rather than uniformly enforcing stronger geometric smoothness. Height-layer ambiguity is most evident near building boundaries, roof--ground transitions, and other height discontinuities, where vertically separated Gaussian responses may jointly contribute to the same rendered location. The post-hoc replacement experiment shows that simply replacing the EOGS elevation with the selected dominant-layer height provides only limited improvement, whereas optimizing the Gaussian representation with HLC leads to substantially better DSM accuracy. Similarly, direct height-dispersion regularization does not reproduce the improvement of HLC-GS. These observations indicate that the main benefit comes from reliability-guided regulation of competing height hypotheses during optimization rather than from an alternative elevation readout or generic suppression of vertical dispersion.

The diagnostic analyses further clarify when this regulation is most effective. Regions assigned higher risk before HLC optimization generally obtain larger subsequent improvements, and the error-stratified results show that the gains are concentrated in moderate- and high-error regions, with the largest improvement occurring in the highest-error quartile. In contrast, the lowest-error quartile exhibits a small degradation, indicating that the current risk estimation can still trigger unnecessary correction in regions that are already reconstructed accurately. This behavior also explains why the improvement is more pronounced under the stricter PAG$_{2.5}$ criterion, while changes under the more relaxed PAG$_{7.5}$ criterion are comparatively small and may vary across scenes. These results suggest that HLC-GS is most useful when multiple plausible height responses coexist, but less beneficial when the underlying Gaussian geometry already provides a confident and accurate surface estimate.

Several limitations remain. HLC-GS operates on height hypotheses already represented by the Gaussian scene and therefore cannot recover surfaces that are absent because of severe occlusion, insufficient observations, or inaccurate Gaussian geometry. Its effectiveness consequently remains dependent on the quality of the underlying representation. In addition, the degradation in low-error regions indicates that the current reliability estimation does not completely prevent over-correction. Future work will investigate more adaptive uncertainty estimation and visibility-aware reasoning to better distinguish genuinely ambiguous regions from already reliable geometry, while reducing unnecessary intervention in accurately reconstructed areas.

\section{Conclusion}
\label{sec:conclusion}

In this paper, we formulated multi-view satellite 3D reconstruction as a reliability-aware fusion problem over competing height hypotheses within a shared Gaussian representation and introduced HLC-GS, a reliability-aware Height-Layer Consistency Gaussian Splatting framework. HLC-GS addresses geometric ambiguity caused by vertically separated Gaussian responses by organizing them into candidate height hypotheses, evaluating their relative support and reliability, and selectively regulating unreliable dominant and interfering secondary responses during optimization. In this way, reliability-aware fusion is realized through optimization of the shared Gaussian representation, reducing non-physical intermediate elevations caused by conventional alpha-weighted aggregation and improving the recovered surface geometry.

Experiments on seven benchmark scenes from the DFC2019 and IARPA2016 datasets demonstrate the effectiveness of HLC-GS. Compared with the EOGS backbone, HLC-GS reduces the average DSM MAE from \(1.46\,\mathrm{m}\) to \(1.18\,\mathrm{m}\) and RMSE from \(2.78\,\mathrm{m}\) to \(2.58\,\mathrm{m}\), corresponding to relative reductions of \(19.18\%\) and \(7.19\%\), respectively, while increasing PAG$_{2.5}$ from \(86.09\%\) to \(88.61\%\). Diagnostic analyses further show that the improvements are concentrated in ambiguous and higher-error regions, while the same-backbone height-dispersion comparison indicates that the gain cannot be explained solely by generic suppression of vertical dispersion. These results support reliability-aware regulation of competing height hypotheses as an effective strategy for satellite DSM reconstruction. Future work will investigate more adaptive uncertainty estimation, improved visibility-aware reasoning, and the incorporation of land-cover and temporal information for more diverse satellite reconstruction scenarios.


\section*{Declaration of competing interest}

The authors declare that they have no known competing financial interests or personal relationships that could have appeared to influence the work reported in this paper.


\section*{Acknowledgements}

This work was supported by the National Science Fund for Distinguished Young Scholars grant number 62425102, Hubei Province Strategic Talent Cultivation Project No. 2024DJA035, and LIESMARS Special Research Funding.


\bibliographystyle{elsarticle-num}
\bibliography{cas-refs_doi}

\end{document}